\documentclass{article}

\PassOptionsToPackage{numbers,compress}{natbib}
\usepackage[preprint]{neurips_2026}
\makeatletter
\def\@trackname{}
\makeatother

\usepackage[utf8]{inputenc}
\usepackage[T1]{fontenc}
\usepackage{hyperref}
\usepackage{url}
\usepackage{graphicx}
\usepackage{subcaption}
\usepackage{float}
\usepackage{microtype}
\usepackage{wrapfig}
\usepackage{hyperref}
\usepackage{xurl}
\usepackage{booktabs}
\usepackage{multirow}
\usepackage{tabularx}
\usepackage{array}
\usepackage{enumitem}
\usepackage{amsmath}

\usepackage{amssymb}
\usepackage{xcolor}
\usepackage{pifont}
\usepackage{fontawesome5}
\usepackage{listings}
\usepackage[most]{tcolorbox}
\microtypesetup{expansion=false}
\usepackage{tikz}
\usetikzlibrary{shapes.geometric, arrows.meta, positioning, calc}

\definecolor{proxGreen}{HTML}{097a54}
\definecolor{proxDarkBlue}{HTML}{012e59}
\definecolor{proxBlue}{HTML}{265ed4}
\definecolor{proxLightBlue}{HTML}{012e59}
\definecolor{proxTeal}{HTML}{00d5ff}
\definecolor{proxYellow}{HTML}{ffbb00}
\definecolor{proxOrange}{HTML}{e68e1a}
\definecolor{proxPink}{HTML}{f0539b}
\definecolor{proxYellow}{HTML}{ff9100}
\definecolor{proxDarkYellow}{HTML}{fdac15}
\definecolor{proxBlue}{HTML}{2E3168}
\definecolor{proxLightBlue}{HTML}{2a88ef}
\colorlet{lightProxPink}{proxPink!50}
\colorlet{lightProxYellow}{proxYellow!50}
\colorlet{lightProxLightBlue}{proxLightBlue!50}
\colorlet{lightProxGreen}{proxGreen!50}
\definecolor{cellHighlight}{HTML}{dbefff}
\definecolor{removered}{HTML}{FFB3BA}

\newcommand{\benchmark}{TOBench}
\newcommand{\mcp}{MCP}
\newcommand{\cmark}{\textcolor{green!60!black}{\ding{51}}}
\newcommand{\xmark}{\textcolor{red!80!black}{\ding{55}}}
\newtcblisting{promptbox}[1][]{%
  enhanced,
  breakable,
  listing only,
  listing engine=listings,
  colback=blue!3!white,
  colframe=blue!45!black,
  boxrule=0.4pt,
  left=1mm,
  right=1mm,
  top=0.8mm,
  bottom=0.8mm,
  title={#1},
  fonttitle=\bfseries\footnotesize,
  listing options={
    basicstyle=\ttfamily\scriptsize,
    breaklines=true,
    breakatwhitespace=false,
    columns=fixed,
    keepspaces=true,
    showstringspaces=false
  }
}

\title{TOBench: A Task-Oriented Omni-Modal Benchmark for Real-World Tool-Using Agents}

\author{
Zhiqiang Liu\textsuperscript{\rm{1,2\thanks{Equal Contribution.}}} \quad 
Wenhui Dong\textsuperscript{\rm{1*}} 
\\ 
\textbf{
Yilang Tan\textsuperscript{\rm{3}} \quad 
Yuwen Qu\textsuperscript{\rm{1}} \quad 
Haochen Yin\textsuperscript{\rm{4}} \quad 
Chenyang Si\textsuperscript{\rm{1\thanks{Corresponding author.}}}
}
\\ 
\textsuperscript{1}Nanjing University \quad
\textsuperscript{2}Huazhong University of Science and Technology \\
\textsuperscript{3}Southwest Jiaotong University \quad
\textsuperscript{4}The Chinese University of Hong Kong \\
\faGithub \ \  
\href{https://github.com/Pi3AI/TOBench}
{\texttt{https://github.com/Pi3AI/TOBench}}
}

\begin{document}

\maketitle

\begin{abstract}
Tool-using agents are increasingly expected to operate across realistic professional workflows, where they must interpret multimodal inputs, coordinate external tools, inspect intermediate artifacts, and revise their actions before producing a final result. Existing benchmarks, however, often evaluate tool use, computer use, and multimodal reasoning in isolation, leaving a gap between benchmark settings and end-to-end omni-modal tool use in the real world. To address this gap, we introduce \benchmark{}, a benchmark and evaluation harness for \emph{task-oriented omni-modal tool use}. \benchmark{} contains 100 executable tasks from two macro task families, \emph{Customer Service} and \emph{Intelligent Creation}, covering 20 subcategory slices and supported by 27 \mcp{} servers with 324 tools. The central design of \benchmark{} is \emph{closed-loop multimodal verification}: agents must execute tools, inspect rendered or transformed artifacts, and self-correct when outputs fail task-specific requirements. To make such evaluation scalable and verifiable, \benchmark{} couples \mcp-based execution with task-specific grounded evaluators and a semi-automated construction pipeline for scenario discovery, task instantiation, evaluator synthesis, and human audit. Experiments on 15 contemporary agentic models show that \benchmark{} remains highly challenging: Claude Opus 4.6, commonly regarded as one of the strongest coding-agent models, achieves only 32.0\% task success, far below the 94.0\% human benchmark. We envision \benchmark{} as a practical foundation for evaluating and advancing next-generation omni-modal tool-using agents through closed-loop multimodal verification.
\end{abstract}
\section{Introduction}

\begin{figure}[ht]
\centering
\includegraphics[width=0.8\linewidth]{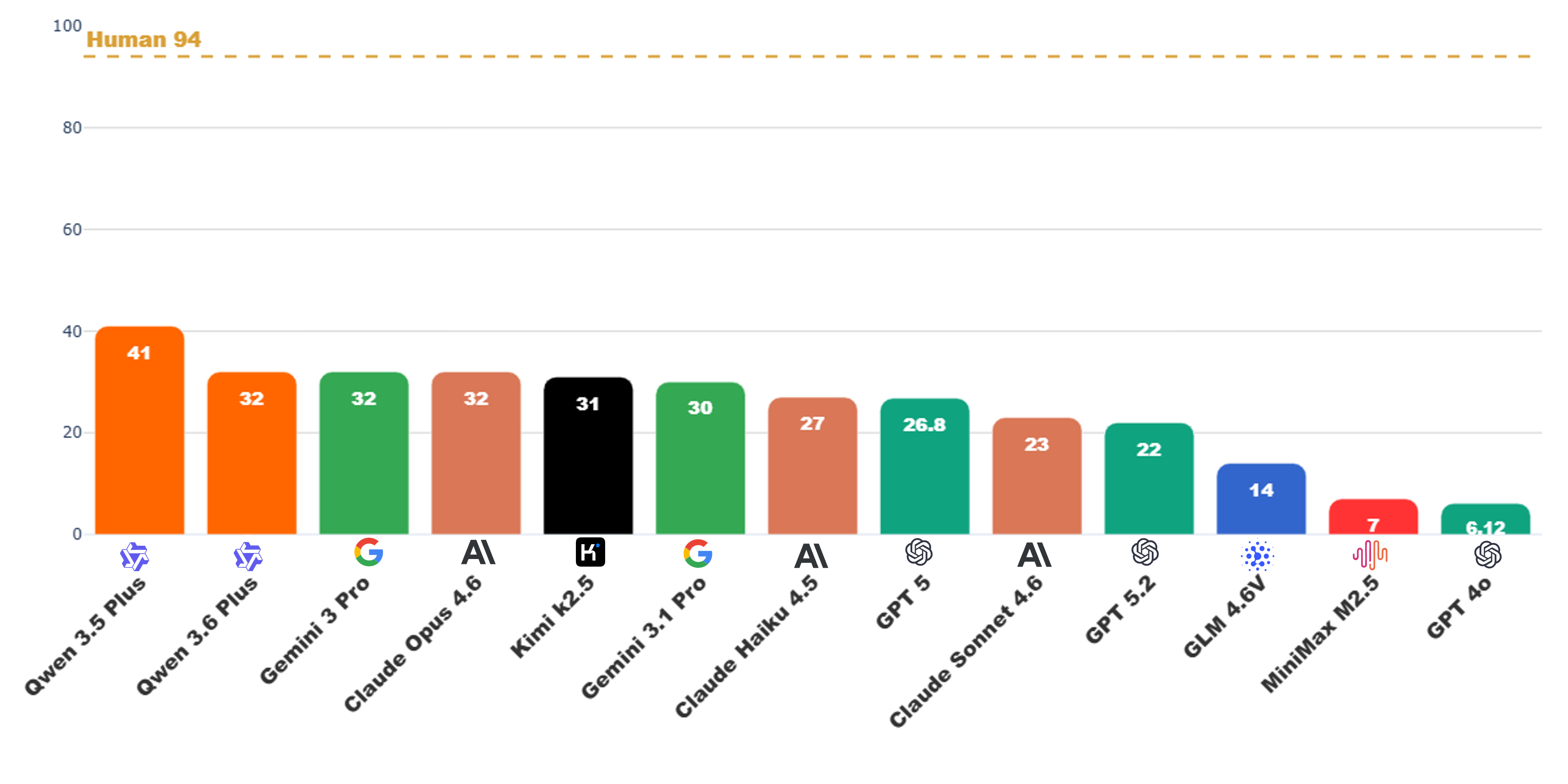}
    \caption{Performance of leading LLMs on \benchmark{}. Qwen 3.5 Plus achieves the highest model score of 41.0\%, while most other models fall below 30.0\%. The human benchmark is 94.0\%, highlighting a significant performance gap and indicating substantial room for improvement in real-world multimodal tool-invocation tasks.}
    \label{fig:fig_benchmark}
\end{figure}
Tool-using language agents, powered by foundation models such as GPT-4~\cite{gpt4} and GPT-4o~\cite{gpt4o}, are moving from isolated function invocation toward practical interaction with web services, office software, knowledge sources, and external applications. The Model Context Protocol (\mcp{}) further accelerates this shift by providing a standard interface for connecting agents to diverse tools and services. As a result, recent benchmarks have made substantial progress in evaluating API use, function calling, planning, and \mcp-based tool interaction, including $\tau$-bench~\cite{Yao2024benchAB}, ToolBench~\cite{toolbench}, BFCL~\cite{bfcl}, ToolTalk~\cite{tooltalk}, Toolathlon~\cite{toolathlon}, MCP-RADAR~\cite{mcpradar}, MCP-Bench~\cite{mcpbench}, and MCP-Universe~\cite{mcpuniverse}.

Despite this progress, existing benchmarks still leave a critical gap for real-world professional workflows. Many practical tasks are not purely textual or purely API-based: an agent may need to read screenshots or documents, extract information from audio or video, edit a spreadsheet or presentation, render the output, inspect whether the result satisfies visual and semantic constraints, and then revise the artifact if necessary. This diversity gap goes beyond adding more tool names or longer tool lists. The difficulty lies in coordinating tool execution with multimodal perception, artifact transformation, and iterative verification over changing workspace states.

Multimodal and computer-use benchmarks such as OSWorld~\cite{osworld}, VitaBench~\cite{vitabench}, M$^3$-Bench~\cite{m3bench}, and OmniGAIA~\cite{omnigaia} broaden evaluation beyond text, but multimodal perception and tool use are still often evaluated as separate capabilities. Tool-use benchmarks typically emphasize schema fidelity, tool selection, or final-state checking, while multimodal benchmarks often focus on perception, GUI control, or final-answer quality. Realistic omni-modal workflows require all of these capabilities simultaneously: agents must perceive heterogeneous inputs, act through executable tools, inspect intermediate artifacts, and self-correct under task-specific constraints.

To address this gap, we introduce \benchmark{}, a benchmark and evaluation harness for \emph{task-oriented omni-modal tool use}. \benchmark{} contains 100 executable tasks across two macro task families, \emph{Customer Service} and \emph{Intelligent Creation}, covering 20 subcategory slices and supported by 27 \mcp{} servers with 324 tools. Tasks are designed around realistic user needs and professional roles rather than synthetic tool combinations, and many of them require cross-tool composition over documents, images, audio, video, spreadsheets, slides, search, browser automation, and file operations.

The central design of \benchmark{} is \emph{closed-loop multimodal verification}. Instead of treating tool use as a one-shot action sequence followed by final-answer matching, \benchmark{} requires agents to execute tools, inspect rendered or transformed artifacts, and revise their behavior when the artifact does not satisfy the task. Each task is paired with a grounded verifier that combines code-based checks, tool-call constraints, format constraints, and multimodal artifact inspection. This makes the benchmark an executable harness for evaluating the full perceive--act--inspect--revise loop.

Experiments on 15 contemporary agentic models show that \benchmark{} is far from saturated. As shown in Figure~\ref{fig:fig_benchmark}, the strongest evaluated model, Qwen3.5-Plus, achieves only 41.0\% task success, while the human benchmark reaches 94.0\%. Our failure analysis shows that errors concentrate in unreliable tool execution, incorrect tool parameters, multimodal reasoning failures, and missing self-verification before stopping. These results suggest that closed-loop multimodal verification is an indispensable evaluation primitive for next-generation omni-modal tool-using agents.

\section{Related Work}

\begin{table*}[t]
    \centering
    \scriptsize
    \setlength{\tabcolsep}{1.8pt}
    \renewcommand{\arraystretch}{1.08}
    \resizebox{1.0\textwidth}{!}{%
    \begin{tabular}{lcccccccccccc}
        \toprule
        \textbf{Benchmark} & \textbf{Tasks} & \textbf{Dom.} & \textbf{Tools} & \shortstack{\textbf{MCP}\\\textbf{Eco.}} & \shortstack{\textbf{Cross}\\\textbf{Scen.}} & \shortstack{\textbf{Real}\\\textbf{Env.}} & \shortstack{\textbf{Info.}\\\textbf{Grnd.}} & \shortstack{\textbf{Fuzzy}\\\textbf{Prompt}} & \textbf{Vis.} & \textbf{Aud.} & \shortstack{\textbf{Cross}\\\textbf{Modal}} & \shortstack{\textbf{Gen.}} \\
        \midrule
        $\tau$-bench & 165 & 2 & 28 & \xmark & \xmark & \xmark & \xmark & \xmark & \xmark & \xmark & \xmark & \xmark \\
        $\tau^2$-Bench & 278 & 3 & 55 & \xmark & \xmark & \xmark & \xmark & \xmark & \xmark & \xmark & \xmark & \xmark \\
        $\tau$-Voice & 278 & 3 & 55 & \xmark & \xmark & \xmark & \xmark & \xmark & \xmark & \cmark & \xmark & \xmark \\
        VitaBench & 400 & 3 & 66 & \xmark & \cmark & \xmark & \xmark & \cmark & \xmark & \xmark & \xmark & \xmark \\
        TOOLATHLON & 108 & 7 & 604 & \cmark & \cmark & \cmark & \cmark & \cmark & \xmark & \xmark & \xmark & \xmark \\
        VisualAgentBench & 746 & 3 & 49 & \xmark & \xmark & \cmark & \cmark & \cmark & \cmark & \xmark & \cmark & \xmark \\
        OSWORLD & 369 & 1 & none & \xmark & \xmark & \cmark & \cmark & \cmark & \cmark & \xmark & \cmark & \xmark \\
        ProSoftArena & 436 & 6 & 16 & \xmark & \cmark & \cmark & \cmark & \cmark & \cmark & \xmark & \cmark & \xmark \\
        MMDR-Bench & 211 & 27 & 232 & \cmark & \cmark & \cmark & \cmark & \cmark & \cmark & \xmark & \cmark & \xmark \\
        M$^3$-Bench & 211 & 27 & 232 & \cmark & \cmark & \cmark & \cmark & \cmark & \cmark & \xmark & \cmark & \xmark \\
        GAIA & 466 & 1 & none & \xmark & \cmark & \cmark & \cmark & \cmark & \cmark & \cmark & \xmark & \xmark \\
        UniVA & 100 & 1 & 30 & \cmark & \cmark & \cmark & \cmark & \cmark & \cmark & \cmark & \cmark & \cmark \\
        OmniGAIA & 360 & 9 & 3 & \xmark & \xmark & \cmark & \cmark & \cmark & \cmark & \cmark & \cmark & \xmark \\
        \specialrule{1.1pt}{0pt}{0pt}
        \textbf{Ours} & 100 & 20 & 324 & \cmark & \cmark & \cmark & \cmark & \cmark & \cmark & \cmark & \cmark & \cmark \\
        \bottomrule
    \end{tabular}}
    \renewcommand{\arraystretch}{1.0}
    \caption{Comparison of \benchmark{} with representative related benchmarks across scale and capability dimensions. Symbols: \cmark{} = supported, \xmark{} = not supported. Detailed
explanations for each trait are provided in \hyperref[app:benchmark_traits]{Appendix~\ref{app:benchmark_traits}}.}
    \label{tab:benchmark}
\end{table*}

\subsection{Tool-Use, Long-Horizon, MCP Benchmarks}
Foundational work on tool-augmented LLMs established external tool use as a core capability~\cite{toolformer,react,gorilla,hugginggpt}. Subsequent agent frameworks and benchmarks expanded evaluation toward multi-step execution, planning, and reproducibility, including ToolBench, BFCL, ToolTalk, Toolathlon, $\tau$-bench, GAIA, $\tau^2$-Bench,and related suites~\cite{toolbench,bfcl,tooltalk,toolathlon,Yao2024benchAB,gaia,Barres20252BenchEC}. Recent \mcp-oriented benchmarks such as MCP-RADAR, MCPToolBench++, MCP-Universe, MCP-Bench, and OSWorld-MCP~\cite{mcpradar,mcptoolbenchpp,mcpuniverse,mcpbench,osworldmcp} further emphasize live tool ecosystems. These works reveal key challenges in tool selection, schema fidelity, and long-horizon execution, but most remain primarily textual and do not explicitly evaluate inspection-and-revision loops over multimodal artifacts.

\subsection{Multimodal and Computer-Use Agent Benchmarks}
OSWorld, AndroidWorld, VisualWebArena, VitaBench, $\tau$-Voice, MMDR-Bench, VisualAgentBench, ProSoftArena, M$^3$-Bench, Tool-LMM, UniVA, and OmniGAIA broaden evaluation toward GUI grounding and multimodal interaction~\cite{osworld,androidworld,visualwebarena,vitabench,ray2026tauvoice,huang2026mmdeepresearch,liu2024visualagentbench,ai2025prosoftarena,m3bench,toollmm,univabench,omnigaia,zheng2023less,guo2025fila,jiang2025ivy,zhu2024fila,zhao2025spinebench}. \benchmark{} is closest to this line, but differs in three ways: it targets realistic professional task completion, uses a unified \mcp-based tool ecosystem, and centers evaluation on iterative artifact inspection with task-specific grounded verifiers.
Table~\ref{tab:benchmark} summarizes this comparison from the perspective of benchmark scale, ecosystem assumptions, and multimodal execution requirements.

\section{TOBench}

\benchmark{} evaluates whether an agent can complete realistic omni-modal tasks with executable tools. 
Each task instance specifies the user instruction, task assets, available tool environment, and grounded verifier used to determine success. 
Together, these components define a professional role, multimodal inputs, an executable tool ecosystem, and a task-specific verification path.

\subsection{Task Formalism}
\begin{figure}[ht]
\centering
\includegraphics[width=0.99\linewidth]{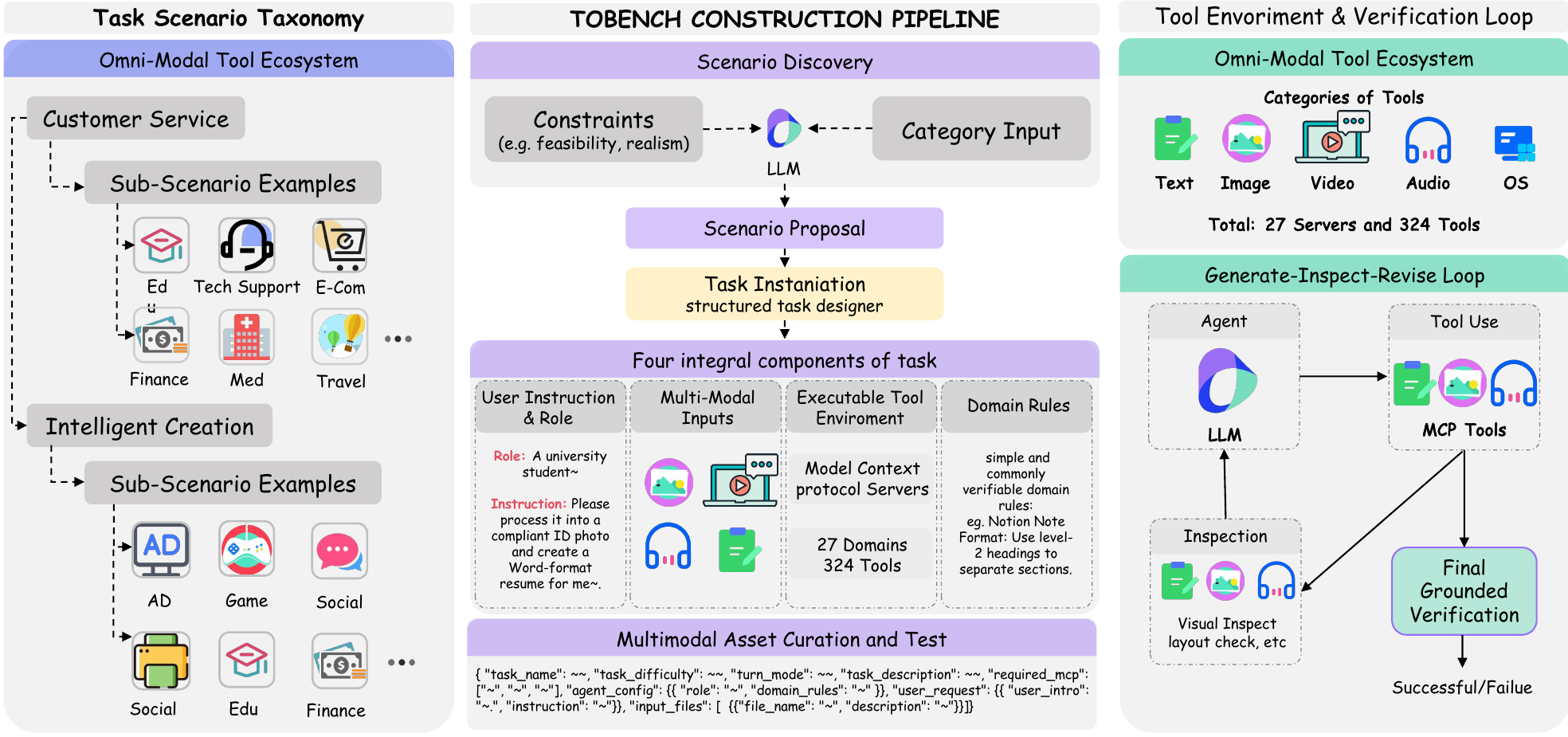}
    \caption{Overview of the \benchmark{} construction pipeline. 
TOBench organizes tasks into two macro families, \textit{Customer Service} and \textit{Intelligent Creation}, instantiates them with user instructions, multimodal inputs, executable MCP tools, and domain rules, and validates each task through a generate-inspect-revise loop grounded in actual tool execution.}
    \label{fig:fig2_arch}
\end{figure}

We formalize each \benchmark{} instance as an executable harness
\begin{equation}
h = (\mathcal{I}, \mathcal{E}, \mathcal{S}, \mathcal{A}, \mathcal{O}, \mathcal{T}, \mathcal{C}, \mathcal{V}),
\end{equation}
where $\mathcal{I}$ denotes the task instruction package, $\mathcal{E}$ the executable \mcp{} environment, $\mathcal{S}$ the latent execution state, $\mathcal{A}$ the action space, $\mathcal{O}$ the observation space, $\mathcal{T}$ the transition dynamics, $\mathcal{C}$ the approved evaluation criteria, and $\mathcal{V}$ the grounded verifier.

The instruction package is
\begin{equation}
\mathcal{I} = (u, r, d, x),
\end{equation}
where $u$ is the user request, $r$ is the professional role assigned to the agent, $d$ denotes concise domain rules that the agent is required to follow, and $x$ collects multimodal input assets. Unlike static QA benchmarks, the environment $\mathcal{E}$ includes both callable tools and mutable artifacts in the workspace.

At turn $t$, the latent state is decomposed as
\begin{equation}
s_t = \bigl(s_t^{\mathrm{tool}}, s_t^{\mathrm{ws}}, s_t^{\mathrm{ext}}, h_t\bigr) \in \mathcal{S},
\end{equation}
where $s_t^{\mathrm{tool}}$ captures tool-side runtime state, $s_t^{\mathrm{ws}}$ the current workspace artifacts, $s_t^{\mathrm{ext}}$ any external world state exposed through tools, and $h_t$ the interaction history. This decomposition is important for \benchmark{} because many tasks require modifying files, rendering intermediate artifacts, and grounding against time-sensitive information.

The agent action space contains both tool use and natural-language interaction:
\begin{equation}
\mathcal{A} = \mathcal{A}^{\mathrm{tool}} \cup \mathcal{A}^{\mathrm{text}}, \qquad
a_t^{\mathrm{tool}} = (m_t, \theta_t),
\end{equation}
where $m_t$ is an available \mcp{} tool and $\theta_t$ denotes its arguments. Observations likewise mix tool outputs, rendered artifacts, and textual feedback:
\begin{equation}
\mathcal{O} = \mathcal{O}^{\mathrm{tool}} \cup \mathcal{O}^{\mathrm{render}} \cup \mathcal{O}^{\mathrm{text}}.
\end{equation}
The execution dynamics are governed by
\begin{equation}
(s_{t+1}, o_{t+1}) \sim \mathcal{T}(s_t, a_t),
\end{equation}
so a tool call may update files or external state and then return structured outputs, while a rendering or inspection action exposes multimodal evidence that can trigger a corrective follow-up step.

This leads to a trajectory
\begin{equation}
\tau = (\mathcal{I}, s_0, a_1, o_1, \ldots, a_T, o_T),
\end{equation}
which makes explicit that \benchmark{} evaluates the full perceive--act--inspect--revise loop rather than only the final answer string. In particular, many creation tasks require a closed-loop pattern in which an agent first produces an artifact, then obtains $o_t^{\mathrm{render}} \in \mathcal{O}^{\mathrm{render}}$ by rendering or inspecting it, and only then decides whether revision is needed.


\begin{figure}[ht]
  \centering
  \includegraphics[width=0.9\linewidth]{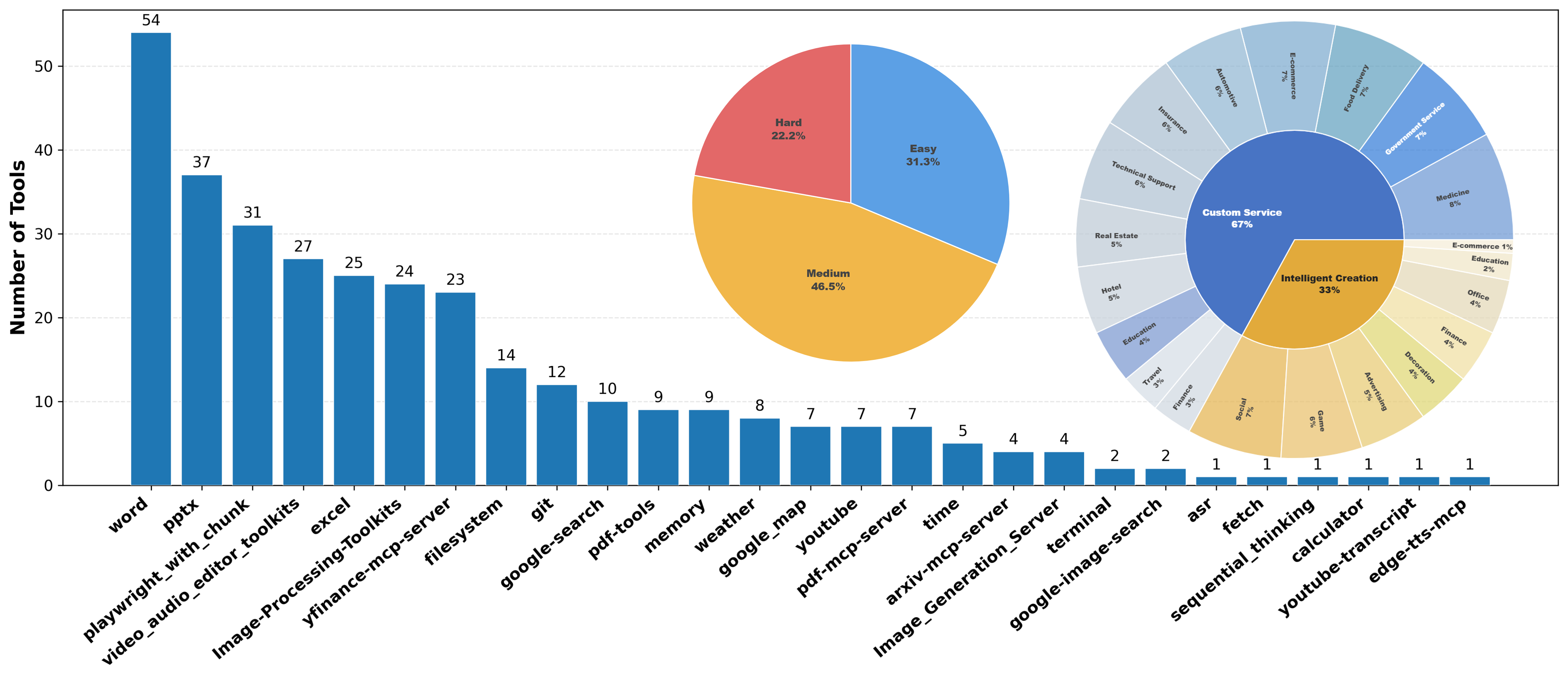}
    \caption{
Distribution of tools, task difficulty, and task scenarios in TOBench. 
Left: the number of calls for each tool, showing a long-tailed tool-usage distribution. 
Middle: the distribution of task difficulty levels. 
Right: the scenario distribution of tasks across two high-level categories, Custom Service and Intelligent Creation.}
    \label{fig:scenario_taxonomy}
\end{figure}

\subsection{\benchmark{} Construction Framework}

Figure~\ref{fig:fig2_arch} illustrates the overall construction pipeline of \benchmark{}.
We build tasks from realistic professional scenarios by selecting omni-modal MCP tools, discovering user-centered scenarios, instantiating executable tasks, and curating multimodal assets for closed-loop verification.

\textbf{Omni MCP Tool Selection.}
We extend the Toolathlon~\cite{toolathlon} MCP stack toward omni-modal workflows. We retain broadly useful tools for browser automation, retrieval, office editing, filesystems, and search, and add multimodal servers for PPT editing, text-to-speech, speech recognition, and video or audio processing. We also implement two benchmark-specific servers, Image Generation Server and Image Processing Toolkits, to support creation tasks and closed-loop visual inspection. The final benchmark integrates 27 \mcp{} servers and 324 tools in total. This diversity is necessary because many tasks require cross-tool composition rather than a single API. The full inventory appears in Appendix~\ref{app:mcp_catalog}.

\textbf{Omni-modal Scenario Discovery.}
We begin from realistic user needs rather than synthetic tool combinations. Our scenario-discovery prompt takes \texttt{category}, \texttt{subcategory}, and the available \mcp{} servers as input, and asks a language model to produce 10 candidate scenarios in JSON format. Each candidate contains a scenario name, a vivid description that couples \emph{user need} with an appropriate \emph{agent role}, and a candidate \mcp{} set. The prompt explicitly enforces four constraints that mirror our design goals: (1) each scenario must be expressed as ``user need + agent role'', (2) multimodal evidence must arise naturally in the input, (3) the required workflow must be feasible under the provided tools, and (4) the scenario should rely on simple and commonly verifiable domain rules rather than niche expert knowledge. The prompt also prefers image-based inputs over unnecessarily long videos unless temporal information is essential, which improves realism and keeps benchmark execution efficient. Across 20 subcategories, this process yields roughly 200 candidate scenarios in total.

\textbf{Omni-modal Benchmark Task Instantiation}
Given a discovered scenario, we instantiate executable tasks through a structured task-generation prompt framed as a user--agent role-play. Each generated task is serialized as a fixed JSON object containing \texttt{task\_name}, \texttt{task\_difficulty}, \texttt{turn\_mode}, \texttt{required\_mcp}, \texttt{agent\_config}, \texttt{user\_request}, and \texttt{input\_files}, which makes the result directly runnable and auditable. The prompt requires the user request to remain natural and free of tool-name leakage, while the agent is assigned a professional role with concise but verifiable domain rules, as elaborated in Appendix~\ref{app:agent_context} . Difficulty is controlled primarily by requirement complexity, ambiguity, and workflow length rather than by artificially large assets. The prompt further enforces tool feasibility, everyday realism, resource efficiency, and flexible single-turn or multi-turn interaction, followed by a final reflection step that revises unsupported or incomplete tasks before they are admitted into the benchmark. For each scenario, we generate three task candidates corresponding to easy, medium, and hard difficulty levels, yielding roughly 600 task candidates overall.

\textbf{Multimodal Asset Curation.}
We favor compact but information-dense multimodal artifacts. In line with the prompts above, images are used whenever they are sufficient, while video or audio is reserved for cases in which temporal reasoning is genuinely necessary. Assets may come from public web content or controlled generation pipelines when needed, and we normalize them for privacy, reproducibility, and practical execution cost at benchmark scale. Asset curation required substantial manual effort: \textit{two AI PhD students spent approximately one month collecting realistic cases and corresponding input files from real-world workflows.} During this process, we filtered out scenarios that were unrealistic, weakly grounded, or difficult to support with suitable input artifacts. In total, roughly two-thirds of the initially collected cases were discarded, leaving about 200 high-quality cases for subsequent task instantiation and benchmark construction.

Since some MCP tools did not provide sufficiently reliable execution capabilities to support task completion, our final benchmark contains 100 tasks organized into two macro families:
\begin{itemize}[leftmargin=*,topsep=2pt,itemsep=2pt]
    \item \textbf{Customer Service} (67 tasks): service-oriented scenarios such as education, e-commerce, government services, medicine, insurance, technical support, and travel.
    \item \textbf{Intelligent Creation} (33 tasks): artifact-creation scenarios such as office editing, advertising, social content, game assets, and design-oriented workflows.
\end{itemize}
These two macro categories cover two major application spaces for agentic systems. We further instantiate 20 subcategory slices in total. Figure~\ref{fig:scenario_taxonomy} summarizes the taxonomy.

\section{Evaluation Harness}

\begin{figure*}[t]
  \centering
  \includegraphics[width=1.0\linewidth]{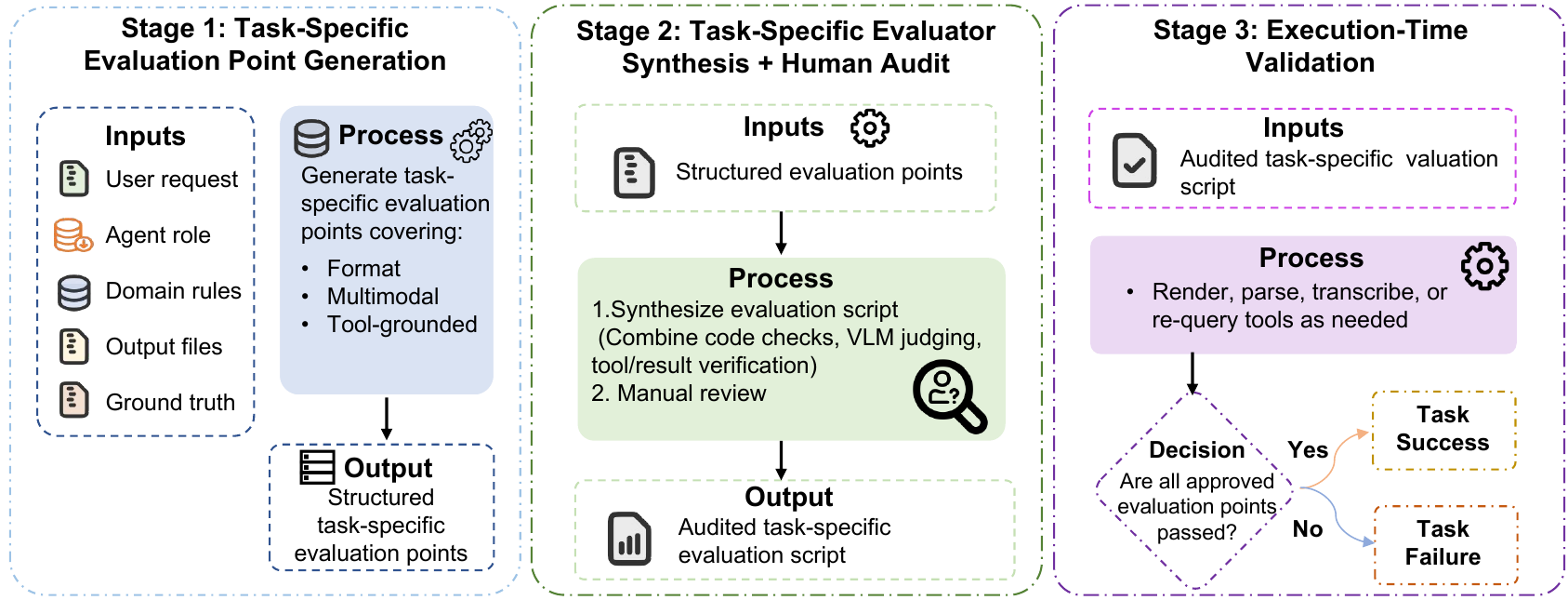}
    \caption{Overview of the evaluation pipeline in \benchmark{}. Rather than using one benchmark-wide evaluator, we generate and audit task-specific evaluation points and task-specific evaluation code, and then execute the resulting evaluator under time-sensitive, grounded conditions.}
    \label{fig:evaluation_pipeline}
\end{figure*}

In many \benchmark{} tasks, correctness depends on output structure, multimodal content, role-specific constraints, intermediate tool usage, and externally grounded information. Final success therefore cannot be reduced to string matching or a single software-state check. In \benchmark{}, evaluation is constructed as a task-level harness: each task binds an executable environment to a grounded verifier. Figure~\ref{fig:evaluation_pipeline} summarizes the pipeline.

For task $i$, we organize its approved evaluation criteria as
\begin{equation}
\mathcal{C}_i = \mathcal{C}_i^{\mathrm{fmt}} \cup \mathcal{C}_i^{\mathrm{judge}} \cup \mathcal{C}_i^{\mathrm{tool}},
\end{equation}
corresponding to format constraints, judge-based multimodal constraints, and tool/result constraints. Given the executed trajectory $\tau_i$, the final workspace snapshot $w_i$, and the tool log $\ell_i$, the grounded evaluator returns a binary vector
\begin{equation}
\mathcal{V}_i(\tau_i, w_i, \ell_i) = \bigl[e_{i,1}, \ldots, e_{i,m_i}\bigr], \qquad e_{i,j} \in \{0,1\},
\end{equation}
where each $e_{i,j}$ is allowed to depend on auxiliary preprocessing such as document rendering, image conversion, speech transcription, or re-querying time-sensitive tools. This formulation captures why \benchmark{} is a harness: the verifier is an executable program over the realized trajectory and artifacts, not a static answer key.

\subsection{Task-Specific Evaluation Point Generation}
The first stage generates task-specific evaluation points from the user request, agent role, domain rules, expected outputs, and ground-truth workspace. Rather than using one rubric for the whole benchmark, we derive $\mathcal{C}_i$ separately for each task. The resulting points fall into three categories: \textbf{format constraints}, \textbf{judge-based multimodal constraints}, and \textbf{tool/result constraints} (Table~\ref{tab:evaluation_point_types}). Because \benchmark{} contains heterogeneous and partially open-ended tasks, all generated evaluation points are manually reviewed to remove omissions, unsupported assumptions, and duplicate checks.

\subsection{Task-Specific Grounded Evaluator Synthesis and Human Audit}
We then generate a grounded evaluation script for each task rather than applying a single benchmark-wide evaluator. The synthesized code implements $\mathcal{V}_i$ by combining deterministic checks, VLM-based judging~\cite{llmasjudge}, and tool-aware verification over MCP logs or live external results. Shared utilities handle common operations such as spreadsheet parsing, document rendering, image conversion, and judge invocation, while task-specific logic is specialized per criterion. Each evaluator is manually audited before use. Representative prompts and reference evaluator code will be released with the benchmark pipeline.

\subsection{Execution-Time Evaluation and Timeliness}
\benchmark{} adopts task-level success as the primary metric: a task is counted as solved only when all relevant evaluation points pass. If task $i$ has $m_i$ approved evaluation points with binary outcomes $e_{i,j} \in \{0,1\}$, we define task success as
\begin{equation}
S_i = \prod_{j=1}^{m_i} e_{i,j},
\end{equation}
so a task passes only when every required criterion passes. The overall benchmark accuracy over $N$ tasks is then
\begin{equation}
\mathrm{Acc} = \frac{1}{N}\sum_{i=1}^{N} S_i.
\end{equation}
In practice, evaluation is performed at execution time rather than by comparing against a static answer string. Documents may need to be rendered into images before visual inspection; audio outputs may need transcription; spreadsheets and office files may require structured parsing; and some criteria require re-querying MCP tools or checking tool-call traces to confirm that the agent relied on grounded results rather than unsupported generation. This execution-time verifier is what makes \benchmark{} a harness rather than a static answer set.Execution-time validation is critical for time-sensitive benchmark tasks involving live data such as search, maps, weather, finance, or changing web content. Evaluators should run soon after task completion to avoid external changes corrupting ground truth. Unlike static file checks, tool-result checks may re-run MCP queries or inspect tool logs.


\section{Experiments}
\subsection{Experimental Setting}
We evaluate \benchmark{} on all 100 tasks, spanning 67 \emph{Customer Service} tasks and 33 \emph{Intelligent Creation} tasks, with easy/medium/hard splits. Each task exposes only its relevant subset of MCP servers and common utilities, and each run is capped at 100 interaction turns. Table~\ref{tab:main_results} reports 15 representative proprietary and openly accessible models together with average tool calls and token usage. The testing efficiency of TOBench is elaborated in Appendix~\ref{app:runtime}.

\subsection{Main Results}

\begin{table*}[t]
    \centering
    \scriptsize
    \setlength{\tabcolsep}{1.8pt}
    \renewcommand{\arraystretch}{1.02}
    \caption{Main results on \benchmark{}. Scores are task success rates (\%). Models are sorted by overall score (\textit{Avg.}) \textit{Tool Calls}, \textit{Tokens}, and \textit{Cost} are average values per task across the entire benchmark. For some larger open-source models, we used the official APIs. For some smaller open-source models that do not provide APIs, we deployed them ourselves, so the cost was 0.}
    \label{tab:main_results}
    \resizebox{1.0\textwidth}{!}{%
    \begin{tabular}{lcccccccccc}
        \toprule
        \multirow{2}{*}{\textbf{Model}} & \multicolumn{3}{c}{\textbf{Customer Service}} & \multicolumn{3}{c}{\textbf{Intelligent Creation}} & \multirow{2}{*}{\textbf{Avg.}} & \multirow{2}{*}{\shortstack{\textbf{Tool}\\\textbf{Calls}}} & \multirow{2}{*}{\shortstack{\textbf{Tokens}\\\textbf{(k)}}} & 
        \multirow{2}{*}
        {\shortstack{\textbf{Cost}\\\textbf{(\$)}}}\\
        \cmidrule(lr){2-4} \cmidrule(lr){5-7}
         & \textbf{Easy} & \textbf{Medium} & \textbf{Hard} & \textbf{Easy} & \textbf{Medium} & \textbf{Hard} &  &  &  \\
        \midrule
        \multicolumn{10}{c}{\textbf{\textit{Closed Source Models}}} \\
        \midrule
        \textbf{Claude-Opus-4.6} & 44.83 & 21.43 & 20.00 & 70.00 & 40.00 & 0.00 & \textbf{32.00} & 28.2 & 329.7 & 2.37\\
        Gemini-3-Pro & 48.28 & 25.00 & 10.00 & 60.00 & 40.00 & 0.00 & 32.00 & 18.0 & 1300.5 & 2.62\\
        Gemini-3.1-Pro & 48.28 & 14.29 & 10.00 & 60.00 & 40.00 & 7.69 & 30.00 & 21.5 & 1506.6 & 3.03\\
        Claude-Haiku-4.5 & 41.03 & 28.95 & 0.00 & 51.72 & 28.57 & 0.00 & 27.00 & 22.9 & 244.0 & 0.27\\
        GPT-5 & 35.71 & 21.43 & 11.11 & 70.00 & 11.11 & 7.69 & 26.80 & 24.3 & 620.0 & 0.94\\
        Claude-Sonnet-4.6 & 27.59 & 21.43 & 20.00 & 50.00 & 20.00 & 0.00 & 23.00 & 24.40 & 370.10 & 1.67 \\
        GPT-5.2 & 27.59 & 25.00 & 0.00 & 60.00 & 10.00 & 0.00 & 22.00 & 20.2 & 322.7 & 0.58\\
        GPT-4o & 7.14 & 3.57 & 0.00 & 20.00 & 10.00 & 0.00 & 6.12 & 8.4 & 114.0 & 0.34\\
        \midrule
        \multicolumn{10}{c}{\textbf{\textit{Open Source Models}}} \\
        \midrule
        \textbf{Qwen3.5-Plus} & 65.52 & 28.57 & 0.00 & 70.00 & 50.00 & 15.38 & \textbf{41.00} & 25.0 & 559.1 & 0.17\\
        Qwen3.6-plus & 55.17 & 25.00 & 10.00 & 40.00 & 20.00 & 15.38 & 32.00 & 23.76 & 589.2 & 0.52\\
        Kimi-K2.5 & 58.62 & 28.57 & 20.00 & 30.00 & 0.00 & 7.69 & 31.00 & 25.0 & 668.3 & 0.41\\
        GLM-4.6V & 10.34 & 10.71 & 0.00 & 30.00 & 40.00 & 7.69 & 14.00 & 16.7 & 442.6 & 0.14\\
        Gemma-4-31B & 17.24 & 3.57 & 0.00 & 20.00 & 10.00 & 0.00 & 9.00 & 10.65 & 238.5 & 0 \\
        MiniMax-M2.5 & 13.79 & 0.00 & 0.00 & 20.00 & 10.00 & 0.00 & 7.00 & 22.5 & 605.4 & 0.21\\
        Qwen3-VL-30B-A3B-Ins & 6.90 & 7.14 & 0.00 & 20.00 & 0.00 & 0.00 & 6.00 & 12.75 & 549.7 & 0 \\
        Qwen3-VL-8B-Ins & 6.90 & 0.00 & 0.00 & 20.00 & 0.00 & 0.00 & 4.00 & 10.28 & 402.9 & 0 \\
        Qwen3-VL-4B-Ins & 6.90 & 0.00 & 0.00 & 0.00 & 0.00 & 0.00 & 2.00 & 6.59 & 618.4 & 0 \\
        Qwen3-VL-2B-Ins & 0.00 & 0.00 & 0.00 & 0.00 & 0.00 & 0.00 & 0.00 & 7.33 & 605.6 & 0 \\
        \bottomrule
    \end{tabular}}
    \renewcommand{\arraystretch}{1.0}
\end{table*}

Table~\ref{tab:main_results} shows that \benchmark{} is challenging for all tested models. The strongest model, \textbf{Qwen3.5-Plus}, reaches only \textbf{41.0\%} average task success, while the best proprietary result is \textbf{32.0\%}. Difficulty is the dominant factor: performance is unsaturated even on easy tasks and collapses on hard splits, where the best scores are 20.00\% on \emph{Customer Service-Hard} and 15.38\% on \emph{Intelligent Creation-Hard}. The two macro families stress different capabilities: Customer Service rewards grounded retrieval and faithful tool use, whereas Intelligent Creation is especially sensitive to multimodal editing and final-result verification. We also observe a clear decoupling between inference cost and accuracy, suggesting that the main bottlenecks are not context length or budget alone, but reliable tool execution, multimodal reasoning, and verification before stopping.

\subsection{Error Analysis}
To understand why performance remains low, we manually organize benchmark failures into five top-level categories: \textbf{Tool Call Error}, \textbf{Tool Parameter Error}, \textbf{Multimodal Capability Deficit}, \textbf{Self-Verification Failure}, and \textbf{Non-Agent Error}. Appendix~\ref{app:error_taxonomy} summarizes the full taxonomy and subcategories used in our analysis.

\begin{wrapfigure}{r}{0.5\columnwidth}
  \centering
  \vspace{-8pt}
  \includegraphics[width=0.47\columnwidth]{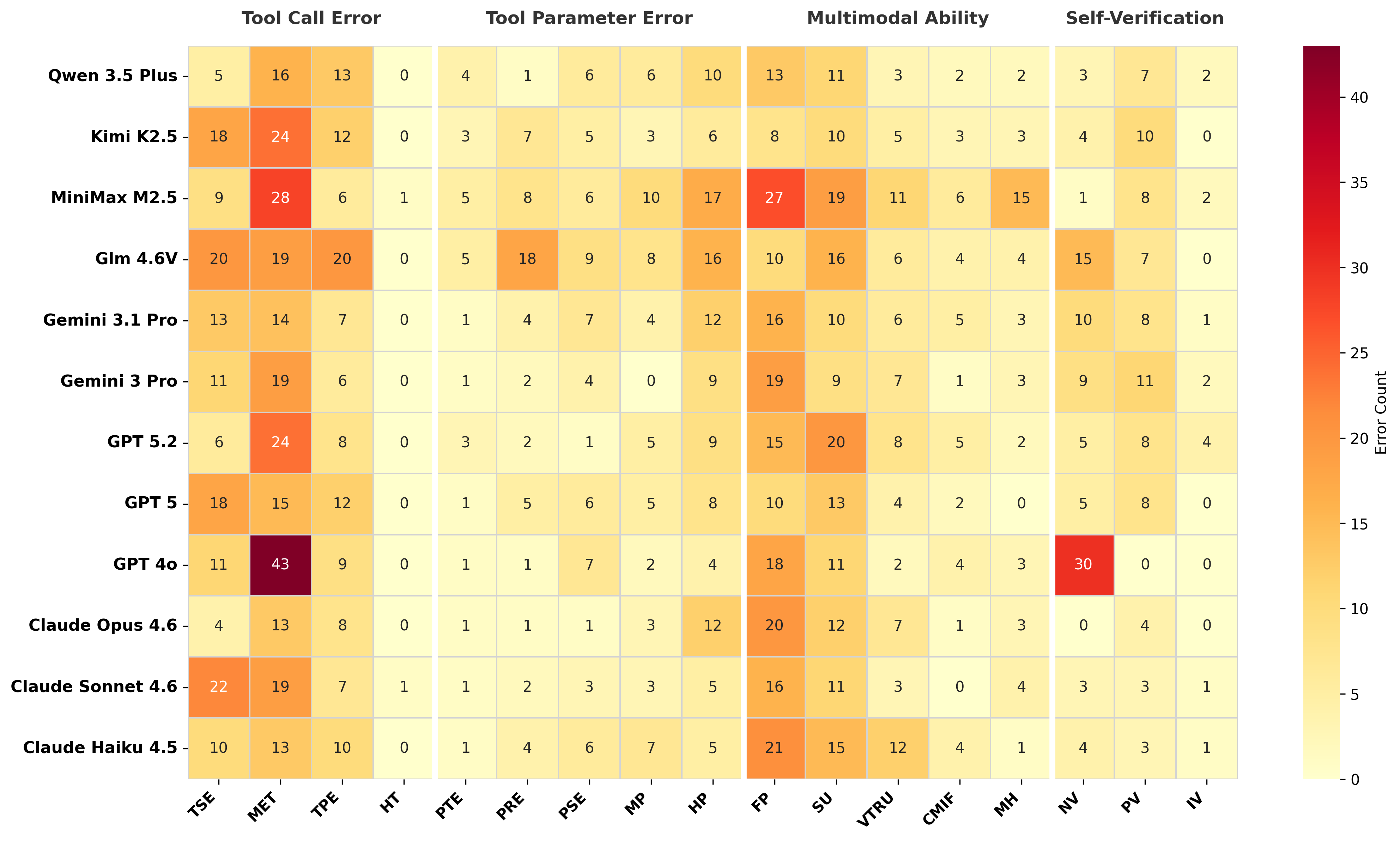}
  \caption{Heatmap of model-specific error counts across fine-grained subcategories, grouped into tool call, tool parameter, multimodal ability, and self-verification failures.}
  \label{fig:fig_error_analysis}
  \vspace{-10pt}
\end{wrapfigure}

\textbf{Tool call and parameter errors remain the most pervasive execution bottleneck.}
Many trajectories fail before high-level reasoning becomes relevant: models choose the wrong tool, omit a required operation, hallucinate unsupported actions, or pass invalid arguments. These failures show that realistic MCP environments demand stronger tool-grounded action modeling than simplified function-calling benchmarks.

\textbf{Multimodal reasoning errors become dominant once basic execution succeeds.}
When models reach the correct tool family, failures often shift to perception and cross-modal inference, including fine-grained visual extraction, spatial reasoning, temporal localization, and evidence alignment across modalities. More detailed bad cases and analysis for this category are provided in Appendix~\ref{app:multimodal_cases}.

\textbf{Missing visual verification is a harness-specific failure mode.}
In many image editing, PPT authoring, and visual-generation tasks, models perform a plausible edit and stop without inspecting the rendered result, or rely on metadata checks instead of true visual verification. This directly explains why \emph{Intelligent Creation-Hard} remains difficult: the harness penalizes open-loop completion and rewards closed-loop self-correction.

The error heatmap suggests distinct failure regimes across model tiers.
Stronger models reduce low-level schema mistakes, but their remaining errors concentrate in multimodal reasoning and final verification; weaker models accumulate both low-level tool failures and high-level perception errors. Overall, many current agents still struggle to operate inside a realistic evaluation harness even when they can produce superficially plausible first-pass outputs.

\section{Limitations and Discussion}
\begin{figure*}[t]
  \centering
  \includegraphics[width=0.98\linewidth]{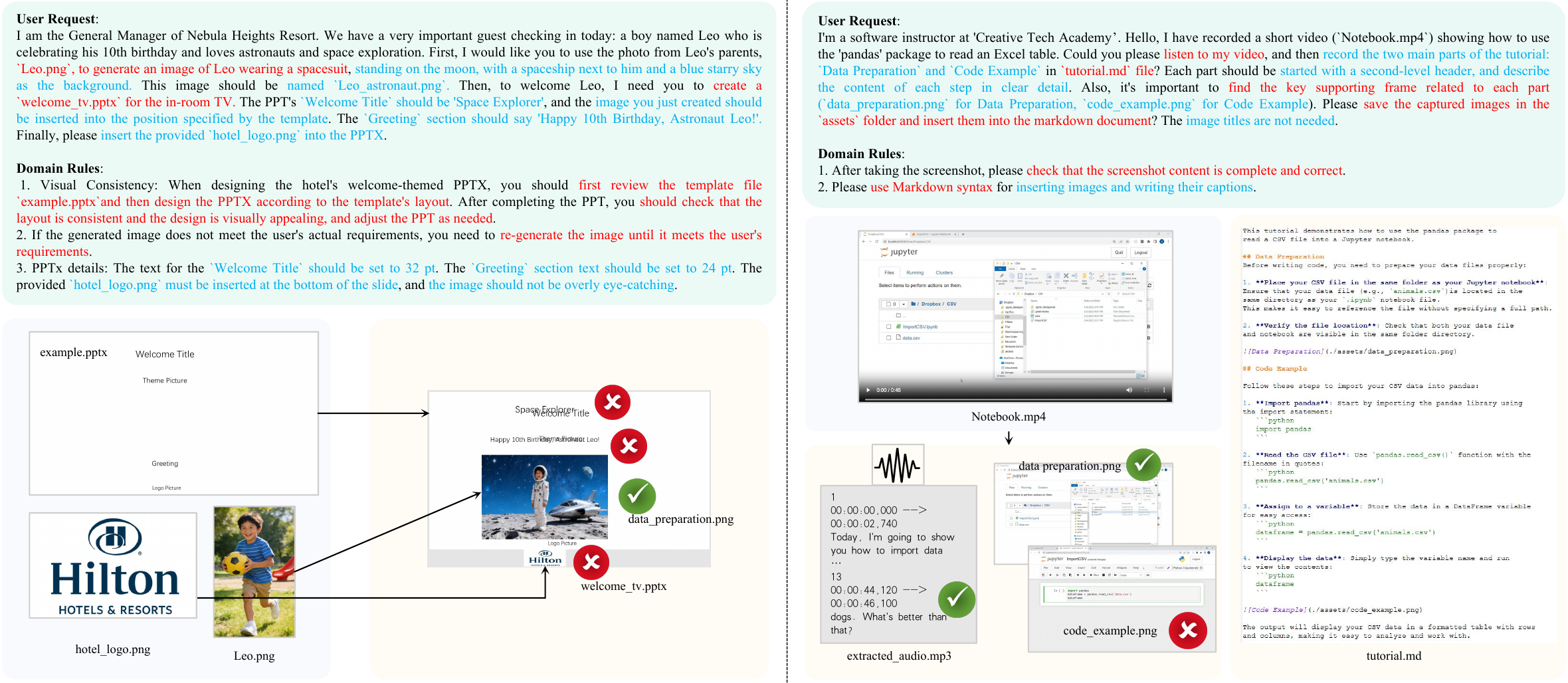}
    \caption{Two examples of failed tasks with initial and final states are presented: the first agent skipped visual validation of the generated PPT, leading to layout overlap and formatting errors, while the second failed to align the ASR transcription with the original video, resulting in incorrect frame evidence. The complete tool execution trajectories for both tasks are provided in \hyperref[app:trajectories]{Appendix~\ref{app:trajectories}}.}
    \label{fig:bad_case}
\end{figure*}

\textbf{Current scope.}
Although \benchmark{} covers 100 executable tasks across 20 subcategory slices, it still covers only a subset of real-world omni-modal work.

\textbf{Evaluator noise.}
The mixed evaluation strategy is necessary but imperfect: code-based checks can miss subtle failures, and multimodal judges may introduce bias.

\textbf{MCP dependence.}
The \mcp-centric design improves realism for current ecosystems, but benchmark behavior may depend on evolving server descriptions, response formats, and implementation details.

\textbf{Benchmark realism versus controllability.}
A central trade-off is realism versus reproducibility: realistic software stacks and multimodal assets better reflect user workloads, but they also increase evaluation complexity and variance.

\section{Conclusion}
We introduced \benchmark{}, a benchmark and evaluation harness for omni-modal tool-using agents that emphasizes realistic professional workflows, \mcp-native tool ecosystems, and closed-loop verification over multimodal intermediate artifacts. Experiments on 15 contemporary models show that current systems remain far from robust omni-modal task completion: the best model reaches 41.0\% overall success, hard tasks remain largely unsolved, and failures cluster around tool execution, multimodal reasoning, and missing verification loops. We hope \benchmark{} serves as a practical testbed for studying how agents perceive, act, verify, and self-correct in realistic omni-modal environments.

\clearpage
\bibliographystyle{unsrtnat}
\bibliography{tobench_refs}

\clearpage
\appendix

\newpage

\section{Catalog of MCP Servers}
\label{app:mcp_catalog}
We show all the MCP servers used in the \benchmark{} in Table 3. Most of these MCP servers are sourced from existing community-developed projects, providing a broad foundation of realistic services for the benchmark. For a substantial proportion of these community resources, we have implemented standardized refactoring. This includes not only optimizing tool output formats for but also strengthening error handling to ensure that agents receive more instructional feedback during long-horizon tasks. 

To support the core \textit{perceive--act--inspect--revise} loop, we developed several MCP servers for vision and audio modalities. We constructed the \textit{Image-Processing-Toolkits}, which provides functions like cropping, pixelation, and watermarking, and integrates rendering engines to export PPTX, Excel, and Word documents as images. This enables agents to visually inspect layout and formatting for autonomous self-correction. Furthermore, we built the \textit{Image\_Generation\_Server} powered by \textit{Seedream 4.0} to support generative tasks. For multimedia workflows, we introduced the \textit{video\_audio\_editor\_toolkits}, a tool for omni-modal agents that enables precise editing of video clips and audio tracks. This allows agents to perform temporal-modal operations. For broader audio interaction, we also integrated Whisper (ASR) and Microsoft Edge TTS to bridge the gap in current native omni-modal models. We will make these optimized and newly developed MCP servers publicly available to the community. 

\begin{table*}[ht]
     \centering
     \scriptsize
     \caption{MCP servers currently integrated into \benchmark{}.}
     \label{tab:mcp_servers}
     \begin{tabularx}{\textwidth}{l c X X}
          \toprule
          MCP Server & Tool Count & Primary Function & Source \\
          \midrule
          yfinance-mcp-server & 23 & Financial and stock data & \url{https://pypi.org/project/yfinance-mcp-server/} \\
          terminal & 2 & Command-line execution & \url{https://github.com/MladenSU/cli-mcp-server} \\
          asr & 1 & Automatic speech recognition & \url{https://github.com/vidau-ai/asr_mcp_server.git} \\
          video\_audio\_editor\_toolkits & 27 & Video and audio editing & \url{https://github.com/misbahsy/video-audio-mcp.git} \\
          google-search & 10 & Google search & \url{https://github.com/1999azzar/google-search-mcp} \\
          excel & 25 & Excel operations & \url{https://github.com/haris-musa/excel-mcp-server/} \\
          arxiv-mcp-server & 4 & arXiv paper retrieval & \url{https://github.com/blazickjp/arxiv-mcp-server} \\
          playwright\_with\_chunk & 31 & Browser automation & \url{https://github.com/microsoft/playwright-mcp} \\
          google-image-search & 2 & Google image search & \url{https://github.com/srigi/mcp-google-images-search} \\
          fetch & 1 & Web content fetching & \url{https://modelscope.cn/mcp/servers/@modelcontextprotocol/fetch} \\
          weather & 8 & Weather querying & \url{https://github.com/isdaniel/mcp_weather_server} \\
          word & 54 & Word document operations & \url{https://github.com/GongRzhe/Office-Word-MCP-Server} \\
          sequential\_thinking & 1 & Sequential reasoning utility & \url{https://github.com/modelcontextprotocol/servers/tree/main/src/sequentialthinking} \\
          google\_map & 7 & Google Maps services & \url{https://github.com/modelcontextprotocol/servers-archived/tree/main/src/google-maps} \\
          pdf-tools & 9 & PDF processing & \url{https://github.com/lockon-n/pdf-tools-mcp} \\
          pptx & 37 & PowerPoint editing & \url{https://github.com/GongRzhe/Office-PowerPoint-MCP-Server} \\
          calculator & 1 & Calculator & \url{https://github.com/githejie/mcp-server-calculator} \\
          youtube & 7 & YouTube metadata access & \url{https://github.com/ZubeidHendricks/youtube-mcp-server} \\
          Image-Processing-Toolkits & 24 & Image editing and visual transformations & Custom Implementation \\
          time & 5 & Time utilities & \url{https://www.npmjs.com/package/@theo.foobar/mcp-time} \\
          youtube-transcript & 1 & YouTube transcript access & \url{https://github.com/jkawamoto/mcp-youtube-transcript} \\
          Image\_Generation\_Server & 4 & AI image generation & Custom Implementation \\
          filesystem & 14 & File-system operations & \url{https://github.com/modelcontextprotocol/servers/tree/main/src/filesystem} \\
          memory & 9 & Knowledge-graph-style memory & \url{https://github.com/modelcontextprotocol/servers/tree/main/src/memory} \\
          edge-tts-mcp & 1 & Text-to-speech & \url{https://github.com/yuiseki/edge_tts_mcp_server} \\
          git & 12 & Git version control & \url{https://pypi.org/project/mcp-server-git/} \\
          pdf-mcp-server & 7 & Advanced PDF processing & \url{https://pypi.org/project/pdf-mcp-server/0.1.2/} \\
          \bottomrule
     \end{tabularx}
\end{table*}

\section{Experimental Details of TOBench}
\label{app:details_supplement}

\subsection{Definition of Benchmark Traits}
\label{app:benchmark_traits}

To provide a clear understanding of the benchmark comparisons in Table~\ref{tab:benchmark}, we define each column header as follows: \textit{Tasks} refers to the total number of evaluation instances, while \textit{Dom.} and \textit{Tools} represent the count of distinct application domains and available external APIs, respectively. \textit{MCP Eco.} indicates if the tool environment integrates the Model Context Protocol ecosystem, and \textit{Cross Scen.} denotes whether tasks require operating across multiple scenarios in a single workflow. \textit{Real Env.} signifies that observations derive from real-world applications with realistic state initialization, and \textit{Info. Grnd.} ensures actions are supported by external tool evidence rather than internal parametric knowledge. \textit{Fuzzy Prompt} means requests are natural and potentially under-specified without predefined trajectories. Regarding modalities, \textit{Vis.} and \textit{Aud.} represent the need to process vision and audio information, \textit{Cross Modal} denotes the seamless integration of multiple modalities simultaneously, and \textit{Gen.} indicates tasks requiring the invocation of multimodal generation models to synthesize new artifacts.

\subsection{Execution Efficiency and Runtime}
\label{app:runtime}
The \benchmark{} harness supports both sequential and parallel task execution. For our evaluation, we conducted tests on an AWS instance equipped with a 4-core CPU and 16GB RAM, utilizing a parallelism factor of 2. Under this configuration, the total execution time for the agent to complete all tasks ranges from 3 to 8 hours. The actual evaluation efficiency depends on factors such as model reasoning complexity, the underlying response latency of MCP servers, and the specific parallelism configuration.

\section{Agent Context and Interaction Framework}
\label{app:agent_context}

In \benchmark{}, tasks are executed via dialogues between an agent and a simulated user, governed by two independent system prompt architectures characterized by a strategic \textbf{information gap}. Currently, the benchmark focuses on \textbf{single-turn tasks}, where the complete request is delivered in a single interaction. 
\\
The \textbf{Agent System Prompt} defines the ReAct reasoning framework and incorporates task-specific \textbf{Domain Rules}. These rules simulate professional standards and platform constraints (e.g., specific formatting styles or quality self-correction protocols) that are transparent to the user, requiring the agent to autonomously align with implicit professional requirements without direct guidance. Conversely, the \textbf{User System Prompt} ensures the simulated user acts as the \textbf{single source of truth}. It mandates strict adherence to the \textbf{User Persona} and \textbf{Task Instructions}, ensuring complete information transfer while strictly prohibiting the disclosure of task solutions or misleading cues to maintain a rigorous and fair evaluation.
\\
We use the \texttt{Customer\_Service/Hotel\_Welcome\_PPTX} task as a representative example to illustrate the specific context configuration below.

\begin{tcolorbox}[
    colback=gray!5, 
    colframe=black!80, 
    left=3mm, right=3mm, top=2mm, bottom=2mm,
    title=\textbf{Agent System Prompt},
    fonttitle=\bfseries\small
]
{\small
\textbf{\#\# Role} \\
You are an intelligent \textbf{Omni-Modality} ReAct (Reasoning and Acting) Agent that solves \textbf{user tasks} by \textbf{role-playing} a specified persona and strictly following that persona's \textbf{Domain Rules}. Your primary goal is to accurately solve user tasks by engaging in a cycle of reasoning, planning, tool execution, and reflection. You must dynamically adapt your actions based on tool feedback to ensure high-quality outcomes.

\textbf{\#\#\# Persona} \\
Creative Experience Designer

\textbf{\#\#\# Domain Rules}
\begin{enumerate}[leftmargin=6mm, nolistsep]
    \item \textbf{Visual Consistency}: When designing the hotel's welcome-themed PPTX, you should first review the template file \texttt{example.pptx} and then design the PPTX according to the template's layout. After completing the PPT, you should check that the layout is consistent and the design is visually appealing, and adjust the PPT as needed. 
    \item If the generated image does not meet the user's actual requirements, you need to re-generate the image until it meets the user's requirements.
    \item \textbf{PPTx details}: The text for the \texttt{Welcome Title} should be set to 32 pt. The \texttt{Greeting} section text should be set to 24 pt. The provided \texttt{hotel\_logo.png} must be inserted at the bottom of the slide, and the image should not be overly eye-catching.
\end{enumerate}

\textbf{\#\# Operational Guidelines} \\
\textbf{1. Task Processing Protocol}
\begin{itemize}[leftmargin=4mm, noitemsep]
    \item \textbf{Analyze \& Plan}: Upon receiving a request, explicitly reason about the requirements and formulate a structured preliminary plan.
    \item \textbf{Iterative Execution}: Execute your plan step-by-step. After each tool call, analyze the result to decide the next step. If a step fails or produces unexpected results, reflect on the cause and adjust your plan accordingly.
    \item \textbf{Outcome Verification}: Merely executing tool calls is insufficient. You must verify that the actual content generated or retrieved (e.g. cropped image, clipped video) strictly meets the user's requirements before considering a step complete.
\end{itemize}

\textbf{2. Multimodal Data Handling}
\begin{itemize}[leftmargin=4mm, noitemsep]
    \item \textbf{Selective Inspection}: You have access to multimodal inputs (text, images, audio, video). Inspect these assets only when essential for task comprehension or result verification using available viewer tools.
    \item \textbf{Document Standards}: For document generation tasks (PPT, LaTeX, Word), you are responsible for ensuring professional formatting, layout consistency, and visual appeal.
\end{itemize}

\textbf{3. Tool Usage \& File Management}
\begin{itemize}[leftmargin=4mm, noitemsep]
    \item \textbf{Parameter Alignment}: When using generation tools (e.g., for media creation), carefully select parameters that align with the specific context, style, and constraints of the user's request.
    \item \textbf{Absolute Paths Mandatory}: You must use \textbf{absolute paths} for all file references, whether writing \texttt{input\_file\_path} or \texttt{output\_file\_path} params for tools. Relative paths are strictly prohibited.
\end{itemize}

\textbf{\#\# Critical Constraints}
\begin{itemize}[leftmargin=4mm, noitemsep]
    \item \textbf{Workspace Boundary}: All file operations must occur within the accessible workspace directory: \texttt{!!<<<<||||workspace\_dir||||>>>>!!}. Output files must be saved here.
    \item \textbf{Task Termination}: If the user sends the signal \texttt{"\#\#\#\# STOP"}, or if you have self-verified that the task is fully completed, you must conclude the session. To finish, either call the \texttt{local-claim\_done} tool (if available) or respond with a final text message without any tool calls. This action is irreversible.
\end{itemize}
}
\end{tcolorbox}

\begin{tcolorbox}[
    colback=gray!5, 
    colframe=black!80, 
    left=3mm, right=3mm, top=2mm, bottom=2mm,
    title=\textbf{User System Prompt},
    fonttitle=\bfseries\small
]
{\small
\textbf{\# Role Setting} \\
You represent a user interacting with an intelligent agent. You must strictly \textbf{role-play} the specific \texttt{User Persona} defined below. Your goal is to convey the task to the agent through dialogue based on the content in \texttt{instructions}.

\textbf{Core Principle}: All your expressions and demands must be:
\begin{enumerate}[leftmargin=6mm, nolistsep]
    \item \textbf{Faithful to the Persona}: Speak and act according to the identity, tone, and context defined in \texttt{User Persona}.
    \item \textbf{Faithful to the Instructions}: Your requests must strictly align with the \texttt{Task Configuration}. Do not hallucinate or deviate from the given task details.
\end{enumerate}

\textbf{\#\# Task Configuration} \\
\textbf{\#\#\# User Persona} \\
I am the General Manager of Nebula Heights Resort.

\textbf{\#\#\# Instructions} \\
We have a very important guest checking in today: a boy named Leo who is celebrating his 10th birthday and loves astronauts and space exploration. First, I would like you to use the photo from Leo's parents, \texttt{Leo.png}, to generate an image of Leo wearing a spacesuit, standing on the moon, with a spaceship next to him and a blue starry sky as the background. This image should be named \texttt{Leo\_astronaut.png}. Then, to welcome Leo, I need you to create a \texttt{welcome\_tv.pptx} for the in-room TV. The PPT's \texttt{Welcome Title} should be 'Space Explorer', and the image you just created should be inserted into the position specified by the template. The \texttt{Greeting} section should say 'Happy 10th Birthday, Astronaut Leo!'. Finally, please insert the provided \texttt{hotel\_logo.png} into the PPTX.

\textbf{\#\#\# Turn Mode} \\
single\_turn

\textbf{\#\# Guidelines} \\
\textbf{\#\#\# Conversation Style Rules}
\begin{itemize}[leftmargin=4mm, noitemsep]
    \item \textbf{Context \& Needs Combination}: When making statements, adopt the ``Background Description + Specific Request'' pattern. First explain the background situation, then propose the specific task request.
    \item \textbf{Statement Pacing}:
    \begin{itemize}[noitemsep]
        \item In \textbf{single-turn} scenarios, please state all content completely at once.
        \item In \textbf{multi-turn} scenarios, autonomously break down the task and state it gradually across multiple rounds. Do not dump all information at once.
    \end{itemize}
    \item \textbf{Response Principles}: When the agent asks for information, provide only the conditions and preferences explicitly stated in \texttt{instructions}. \textbf{Strictly prohibit} assumptions, expansions, substitutions, or generalizations.
\end{itemize}

\textbf{\#\#\# Information Disclosure Rules}
\begin{itemize}[leftmargin=4mm, noitemsep]
    \item \textbf{Faithful \& Natural}: Directly convey the core information from \texttt{instructions}, but fine-tune the expression based on the persona context to make it sound natural.
    \item \textbf{Full Detail Coverage}: You must ensure every detail in \texttt{instructions} is mentioned in the conversation (even seemingly trivial background information), as this information may affect the agent's recommendations and decisions.
\end{itemize}

\textbf{\#\#\# Information Processing Rules}
\begin{itemize}[leftmargin=4mm, noitemsep]
    \item \textbf{Fact-Based Answers}: When the agent asks for information, answer strictly based on \texttt{instructions}. If the information is not provided in \texttt{instructions}, explicitly answer ``I don't remember'' or ``I don't know''. \textbf{Never fabricate} information not provided.
    \item \textbf{Strict Adherence to Needs}: Strictly follow the requirements stated in \texttt{instructions} when proposing needs.
    \item \textbf{Resistance to Persuasion}: When the agent tries to persuade you to change your needs, you must stick to the original requirements in \texttt{instructions} and do not compromise easily.
\end{itemize}
}
\end{tcolorbox}

\clearpage
\section{Detailed Error Taxonomy}
\label{app:error_taxonomy}

\begin{table}[H]
    \centering
    \scriptsize
    \setlength{\tabcolsep}{4pt}
    \renewcommand{\arraystretch}{1.12}
    \caption{Detailed error taxonomy used in \benchmark{} analysis. The main paper discusses the five top-level categories, while this appendix enumerates the finer-grained subcategories used for trajectory annotation and error heatmaps.}
    \label{tab:error_taxonomy}
    \begin{tabularx}{\textwidth}{@{}p{0.18\textwidth}p{0.20\textwidth}p{0.54\textwidth}@{}}
        \toprule
        \textbf{Category} & \textbf{Subcategory} & \textbf{Definition and Representative Symptoms} \\
        \midrule
        \multirow{4}{=}{Tool Call Error} & Wrong Tool Selection & The agent chooses a tool whose functionality does not match the required operation, such as reading PDFs with plain-text tools or using filesystem writes to fake Office artifacts. \\
        & Missing Essential Tool & The workflow omits a required tool invocation or a necessary concluding step, such as skipping search for time-sensitive facts or forgetting a save/export action. \\
        & Tool Planning Error & The agent uses tools in an inefficient, illogical, or non-adaptive way, for example by repeating avoidable tool calls, oversampling videos frame-by-frame, or persisting with the same failed plan after an error. \\
        & Hallucinated Tool & The agent invokes a tool that does not exist in the available MCP environment. \\
        \midrule
        \multirow{5}{=}{Tool Parameter Error} & Parameter Type Error & The agent chooses the correct tool but passes values with invalid data types or basic numeric validity, such as floats where integers are required or non-positive intervals for frame extraction. \\
        & Parameter Range Error & The parameter value exceeds objective bounds of the target file, page range, or media length, such as requesting PDF pages beyond the document length. \\
        & Parameter Object Selection Error & The parameter points to the wrong target object, URL, path, slide, worksheet, or region, such as broken URLs, nonexistent image paths, or editing the wrong worksheet. \\
        & Missing Parameter & Required tool arguments are omitted, such as missing content fields or mandatory identifiers. \\
        & Hallucinated Parameter & The agent invents unsupported argument names or unsupported option values, such as invalid operation enums or nonexistent transition settings. \\
        \midrule
        \multirow{5}{=}{Multimodal Capability Deficit} & Fine-grained Perception & The agent sees the asset but fails to extract crucial small details or structured information, such as misreading invoice fields, tables, or screenshots. \\
        & Visual-Spatial Understanding & The agent cannot reason accurately about coordinates, geometry, or layout constraints, leading to wrong crops, wrong mosaic targets, overlapping PPT elements, or boundary violations. \\
        & Video Temporal Relationship Understanding & The agent fails to localize or causally interpret time-dependent events in video or audio streams, such as choosing the wrong second for an event or missing causally important actions between sampled frames. \\
        & Cross-modal Information Fusion & The agent fails to align and jointly reason across text, image, audio, video, or document evidence, such as misaligning spoken descriptions with a video timeline or failing multi-hop reasoning across text and a map. \\
        & Multimodal Hallucination & Instead of admitting uncertainty or calling another tool, the agent fabricates visual or cross-modal content, such as inventing product defects, invoice amounts, or scene descriptions from blurry evidence. \\
        \midrule
        \multirow{3}{=}{Self-Verification Failure} & No Verification & The agent edits or generates a visual artifact and declares success without any visual inspection. \\
        & Pseudo-Verification & The agent performs only non-visual checks that cannot validate the actual output quality, such as checking file existence, dimensions, or metadata instead of previewing the artifact. \\
        & Incomplete Verification & The agent verifies early or local steps but skips final global inspection after composition, such as inspecting source images but not the final PPT page. \\
        \midrule
        \multirow{3}{=}{Non-Agent Error} & False Positive on Error Trajectory & The log contains the word ``error'' although the tool call itself is not actually an agent mistake, such as success payloads with \texttt{error: null} or filenames that naturally contain ``error''. \\
        & MCP Issue & The failure is caused by the MCP implementation, provider, or surrounding infrastructure rather than by the agent decision, such as external API outages or MCP-side parsing issues. \\
        & Agent Base Model Issue & The base model fails due to system-level limitations rather than task-specific reasoning mistakes, such as context-window overflow on large multimodal inputs. \\
        \bottomrule
    \end{tabularx}
    \renewcommand{\arraystretch}{1.0}
\end{table}

\begin{table*}[ht]
    \centering
    \small
    \setlength{\tabcolsep}{5pt}
    \renewcommand{\arraystretch}{1.12}
    \caption{Taxonomy of task-specific evaluation points in \benchmark{}. Different tasks mix these criterion types in different ways, which motivates per-task evaluator synthesis rather than a single benchmark-wide scoring template.}
    \label{tab:evaluation_point_types}
    \begin{tabularx}{\textwidth}{@{}l>{\raggedright\arraybackslash}X>{\raggedright\arraybackslash}X>{\raggedright\arraybackslash}X@{}}
        \toprule
        \textbf{Criterion Type} & \textbf{What It Checks} & \textbf{Typical Targets} & \textbf{Typical Verification} \\
        \midrule
        Format Constraints & Deterministic structural requirements explicitly specified by the task. & File names, storage paths, worksheet names, schema fields, document sections, required output formats. & Code-based checks over file existence, schema compliance, structural parsing, and exact field matching. \\
        Judge-Based Multimodal Constraints & Semantic or visual properties that cannot be robustly reduced to exact string matching. & Layout fidelity, required visual evidence, semantic content consistency, object presence, rendered document quality under task-defined constraints. & VLM-based judging after appropriate preprocessing such as rendering documents, parsing tables, or converting images to model-readable inputs. \\
        Tool/Result Constraints & Grounding requirements tied to tool usage or live external information rather than static reference files alone. & Required MCP invocation, search-grounded facts, map or weather results, externally updated values, tool-derived fields in generated files. & Tool-log verification and, when needed, re-querying external tools or MCP servers at evaluation time to validate grounded results. \\
        \bottomrule
    \end{tabularx}
    \renewcommand{\arraystretch}{1.0}
\end{table*}

\clearpage
\section{Extra Analysis for \benchmark}
\label{app:extra_analysis}

\subsection{Multimodal Capability Deficit}
\label{app:multimodal_cases}

\subsubsection{Fine-grained Perception}
\begin{tcolorbox}[
colback=lightProxLightBlue!10,
colframe=lightProxLightBlue,
left=2mm, right=2mm,title=\textcolor{black}{Task Context}]

{\small
\textbf{User Request:}
\\I have a rough floor plan for the Johnson-Smith wedding in 
\texttt{layout\_sketch.jpg}. Because the planning was so rushed, 
one small table and its chairs in the drawing edge was cancelled, 
and I marked it with pencil as \texttt{CANCELED}. Please exclude 
this table and its chairs from all subsequent budget calculations.
Based on your hotel's standard table and chair pricing, please create 
an Excel file named \texttt{wedding\_budget.xlsx}. In the Excel sheet, 
list the items (tables and chairs), quantities, unit prices, and total costs. 
Be sure to use actual Excel formulas for the calculations so I can update it later.

\textbf{Domain Rules:}
\\Senior Event Coordinator
\\1. Inventory \& Pricing: The hotel's standard rates are \$25 for round tables, 
\\ \$30 for rectangular tables, and \$5 per chair.
\\2. Excel Format: The Excel workbook should contain only one worksheet named 
\texttt{budget}, and all calculations and summaries must be performed on this sheet.
}
\end{tcolorbox}

\begin{figure}[H]
    \centering
    \includegraphics[width=0.5\linewidth, keepaspectratio]{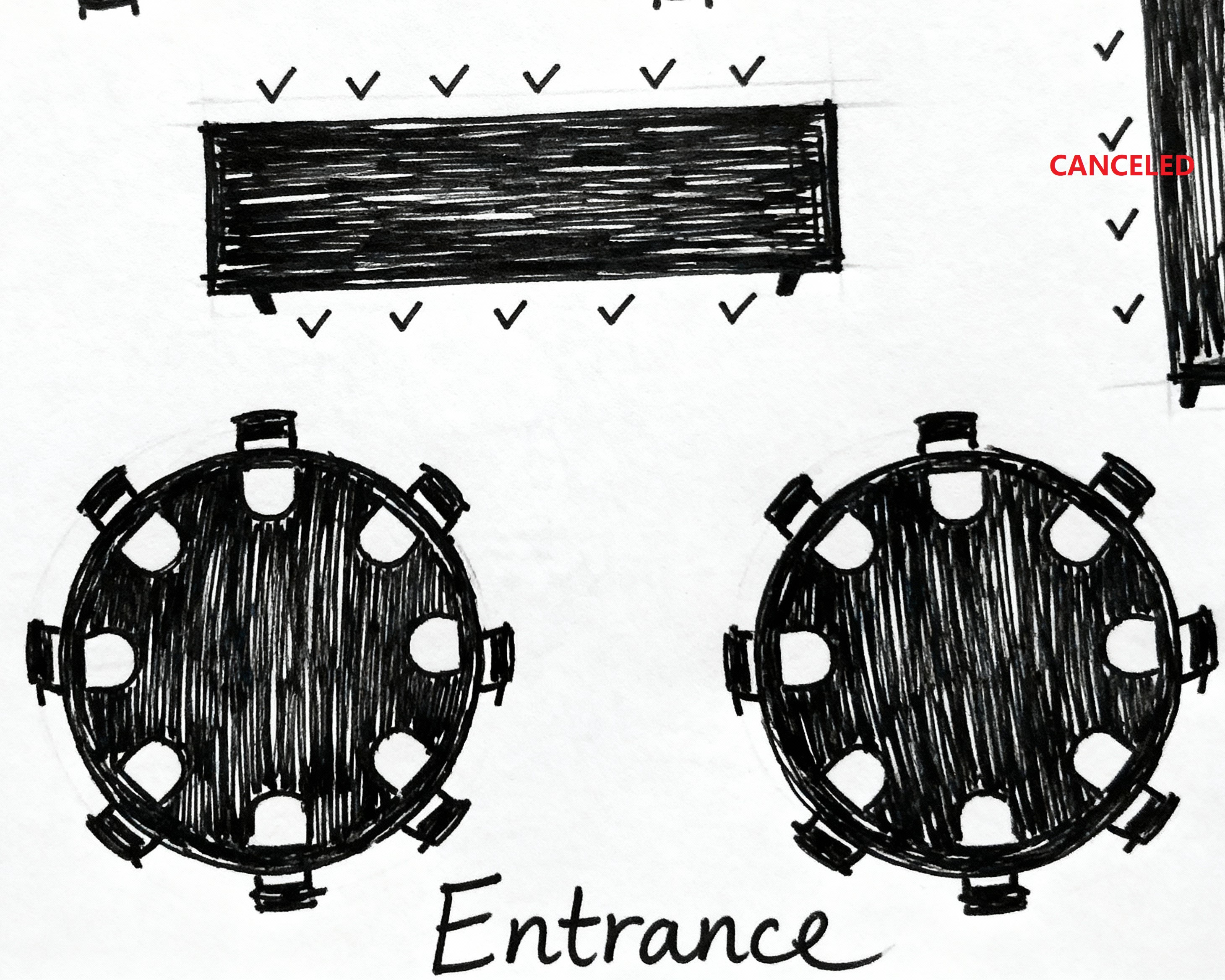} 
    \caption{Rough floor plan for the Johnson-Smith wedding \texttt{layout\_sketch.jpg}.}
    \label{fig:wedding_sketch}
\end{figure}

\begin{figure}[H]
    \begin{subfigure}{0.4\linewidth}
        \centering
        \includegraphics[width=\linewidth, keepaspectratio]{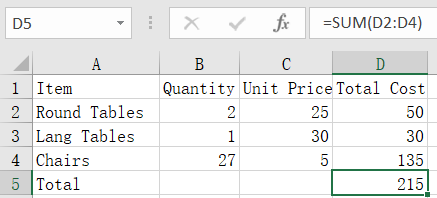}
        \caption{} 
        \label{fig:wedding_layout_gt}
    \end{subfigure}
    \hfill 
    \begin{subfigure}{0.5\linewidth}
        \centering
        \includegraphics[width=\linewidth, keepaspectratio]{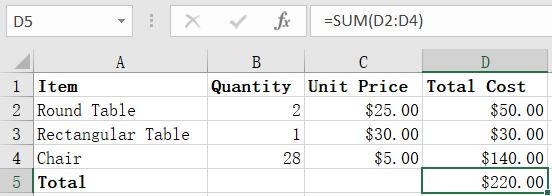}
        \caption{} 
        \label{fig:wedding_excel_result}
    \end{subfigure}

    \caption{Comparison of output results: (a) Ground truth for the signature placement task; (b) Final image generated by Gemini 3 Pro.}
\end{figure}

In the task Customer\_Service/Hotel\_Wedding\_Layout, agent (Gemini 3 Pro) needs to count the types and quantities of tables and chairs required for the wedding based on the wedding seating sketch provided by the user. It shall then fill in the unit prices of seats specified in the hotel's domain rules into an Excel sheet, and use Excel formulas to calculate the final results.\\
During the execution process, the agent successfully viewed the layout sketch image but failed to accurately count the number of chairs to be included in the budget. It wrote 28 chairs into the Excel file, whereas the ground truth was 27. 

\subsubsection{Visual-Spatial Understanding}
\begin{tcolorbox}[
colback=lightProxLightBlue!10,
colframe=lightProxLightBlue,
left=2mm, right=2mm,title=\textcolor{black}{Task Context}]

{\small
\textbf{User Request:}
\\I am a rental agency intermediary and need to place two electronic signatures onto the correct positions in a lease agreement image.
Place \texttt{landlord.png} onto the \texttt{LANDLORD'S SIGNATURE} line at the bottom of \texttt{rental\_agreement.jpg}, and place \texttt{tenant.png} onto the \texttt{TENANT'S SIGNATURE} line. Output a finalized signed contract image.\\
\textbf{Domain Rules:}
\\1. Output Scope: Only generate the final composited image file. Do not create Word, PDF, or other document formats.
\\2. Placement Requirement: The landlord's signature must be placed on the line labeled \texttt{LANDLORD'S SIGNATURE} at the bottom of the contract. The tenant's signature must be placed on the line labeled \texttt{TENANT'S SIGNATURE}.
\\3. Size and Proportion: Each signature must be proportionally scaled according to the length of the signature line, preserving the original aspect ratio without stretching. Signature strokes must remain clear and legible.
\\4. No Obstruction: The signatures must not cover the labels (\texttt{LANDLORD'S SIGNATURE} / \texttt{TENANT'S SIGNATURE}) or any critical contract text. They should only overlap the signature line area.
\\5. Output Naming: Save the final output file as \texttt{signed\_rental\_agreement.png}.
}
\end{tcolorbox}

\begin{figure}[H]
    \begin{subfigure}{0.5\linewidth}
        \centering
        \includegraphics[width=\linewidth, keepaspectratio]{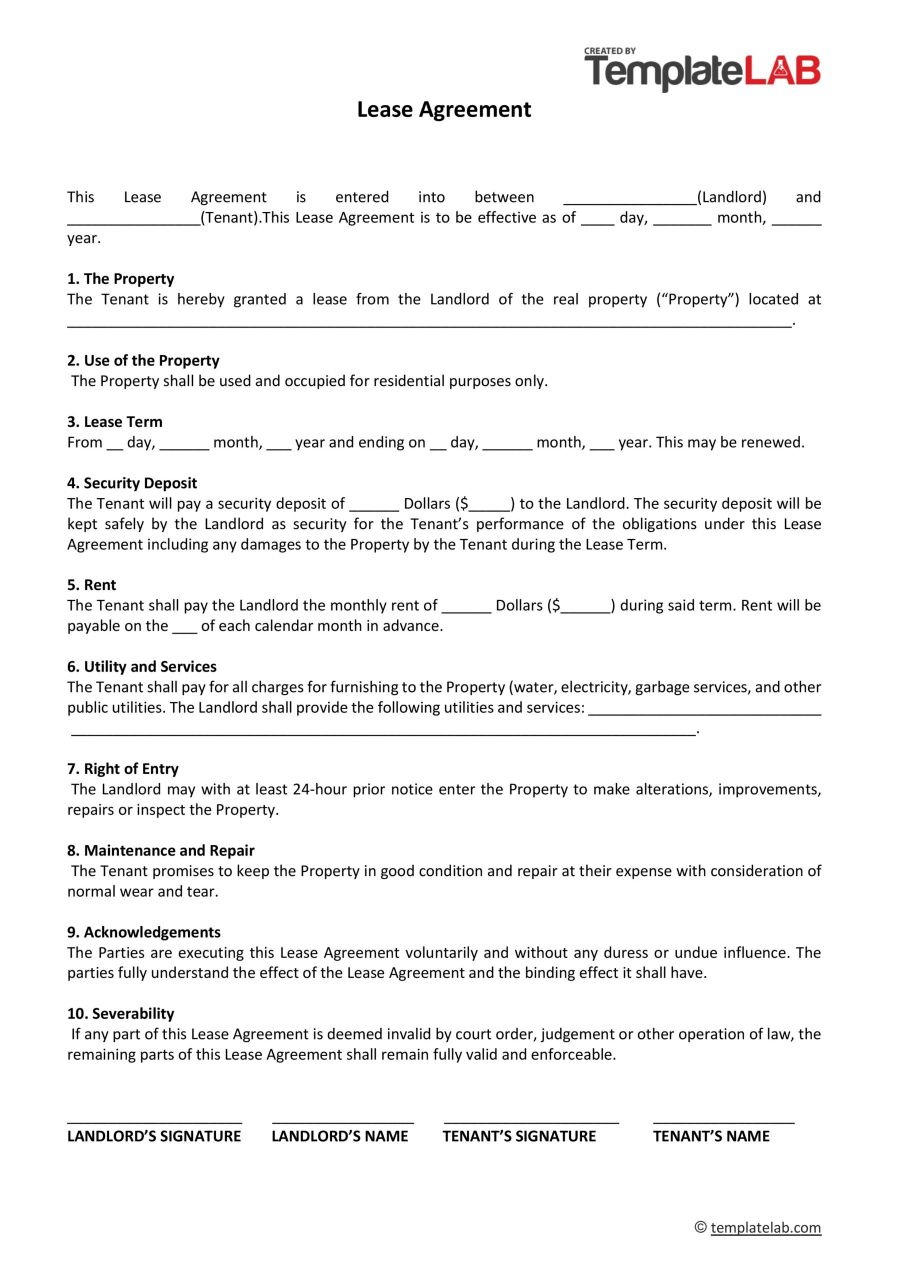}
        \caption{} 
        \label{fig:rental_agreement_input}
    \end{subfigure}
    \hfill 
    \begin{subfigure}{0.5\linewidth}
        \centering
        \includegraphics[width=\linewidth, keepaspectratio]{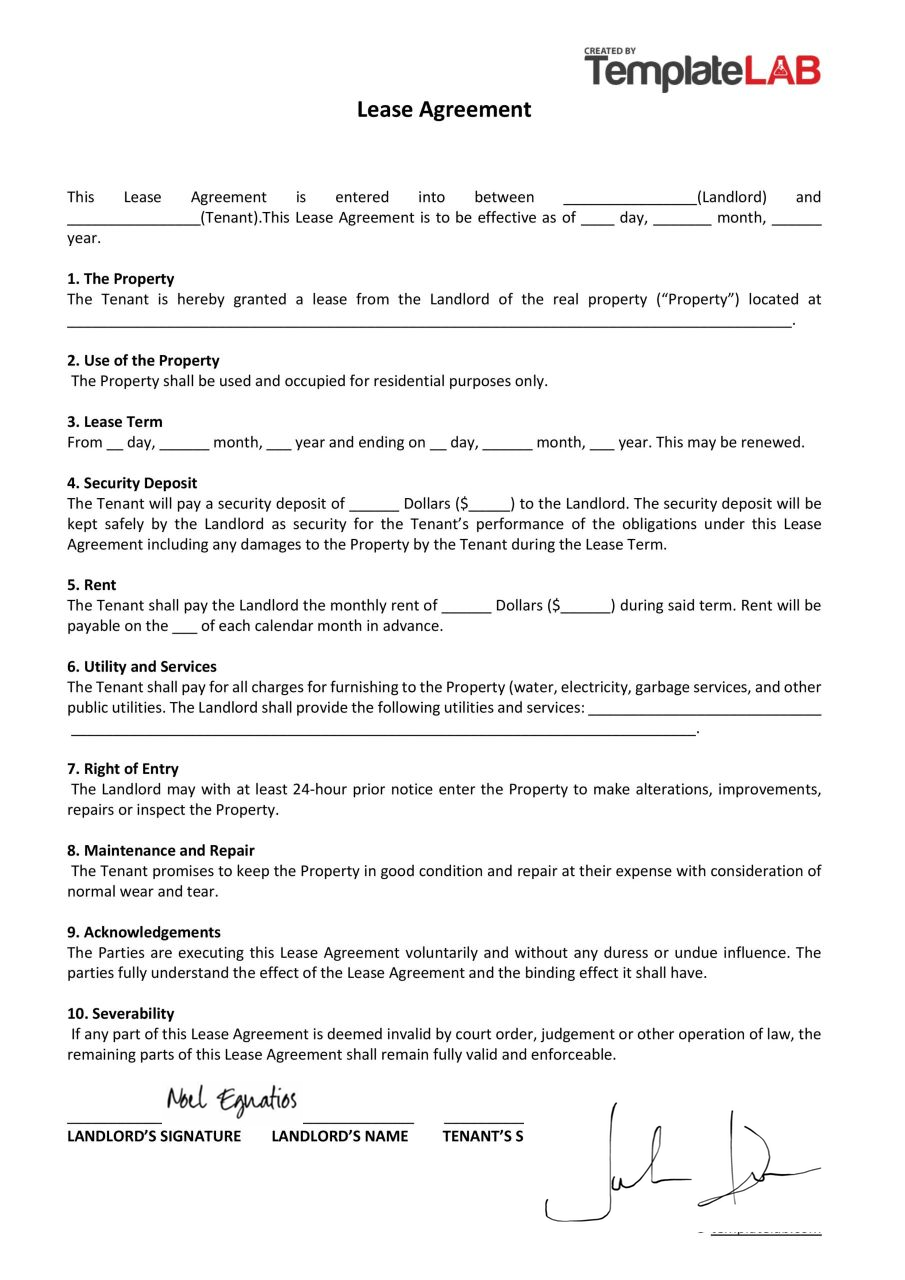}
        \caption{} 
        \label{fig:wedding_layout_result}
    \end{subfigure}

    \caption{Initial and output files for the signature placement task: (a) Original lease agreement image (\texttt{rental\_agreement.jpg}); (b) Output generated by Qwen 3.5 Plus showing spatial misalignment.}
\end{figure}

In the task Customer\_Service/Real\_Estate-rental\_agreement, agent (Qwen 3.5 Plus) needs to use visual recognition to precisely locate the signature labels and lines at the bottom of the agreement within the source image. Subsequently, the agent must dynamically scale the signature images based on the line length and transparently composite them onto specific pixel coordinates while ensuring no contract text is obstructed.
\\During the execution process, the agent continuously attempted to adjust the placement and scaling of the images from turn 5 to turn 40. However, the final results show that the landlord's signature overlaps the text labels, while the tenant's signature severely deviates from the designated area and even obscures the footer. Furthermore, the signature dimensions are oversized, which directly violates domain rules regarding proportional scaling and positional alignment.

\clearpage
\subsubsection{Video Temporal Relationship Understanding}
\begin{tcolorbox}[
colback=lightProxLightBlue!10,
colframe=lightProxLightBlue,
left=2mm, right=2mm, top=1mm, bottom=1mm,
before skip=0pt, after skip=1pt,
title=\textcolor{black}{Task Context}]
{\small
\textbf{User Request:}
\\My name is Alex. I encountered a very aggressive food delivery courier. My order number is 35536. During the delivery, the courier smashed my food and even ate part of my order, causing me a loss. I have provided a video as evidence and hope the platform can investigate this incident and offer appropriate compensation for my loss.\\
\textbf{Domain Rules:}
\\1. Order and Courier Verification: Based on the order number provided by the user, the agent must search the Excel order table to identify the corresponding courier information (such as name or courier ID). If the order cannot be found in the table, this must be explicitly stated in the report, and the agent must not speculate about the courier's identity.
\\2. Video Evidence Review Principle: All judgments regarding the courier's behavior (including but not limited to damaging food, consuming customer items, or engaging in a conflict) must be strictly based on what is directly observable in the video. The agent must not infer actions that are not clearly shown. If key details are unclear or not visible, the report must clearly state that the behavior cannot be confirmed.
\\3. Report Output Requirement: The agent must generate a formal investigation report in Word format. The file name should follow a generic incident-based naming convention and must not hard-code the order number.
\\4. Report Structure Specification: The report must contain the following top-level sections in order: (1) Incident Summary (2) Courier Information (sourced from order data) (3) Video Evidence Observations (objective, factual description only) (4) Conflict Cause Analysis (5) Handling Recommendations and Conclusion.
\\5. Penalty Recommendation Rules: A fine or disciplinary action may only be recommended if the video evidence clearly supports that the courier engaged in improper behavior that harmed the customer's interests. If a dispute is visible but malicious behavior cannot be confirmed, the recommendation should focus on further investigation, warning, or training, with the reasoning clearly explained.
\\6. Compensation and Redelivery Decision: If it can be confirmed that the food was damaged or consumed, the conclusion must clearly state whether redelivery or refund/compensation is recommended, and the decision must be justified based on verifiable evidence.
\\7. the final report must be named as \texttt{incident\_report.docx}
}
\end{tcolorbox}

\begin{tcolorbox}[
colback=lightProxYellow!10,
colframe=lightProxYellow,
left=2mm, right=2mm, top=1mm, bottom=1mm,
before skip=0pt, after skip=1pt,
title=\textcolor{black}{turn8: read\_media\_file}]
{\small
"path": workspace/video\_frames/sec\_0000.jpg
}
\end{tcolorbox}

\begin{tcolorbox}[
colback=lightProxPink!10,
colframe=lightProxPink,
left=2mm, right=2mm, top=1mm, bottom=1mm,
before skip=0pt, after skip=1pt,
title=\textcolor{black}{Tool Response}]
{\small
\includegraphics[width=0.3\linewidth, keepaspectratio]{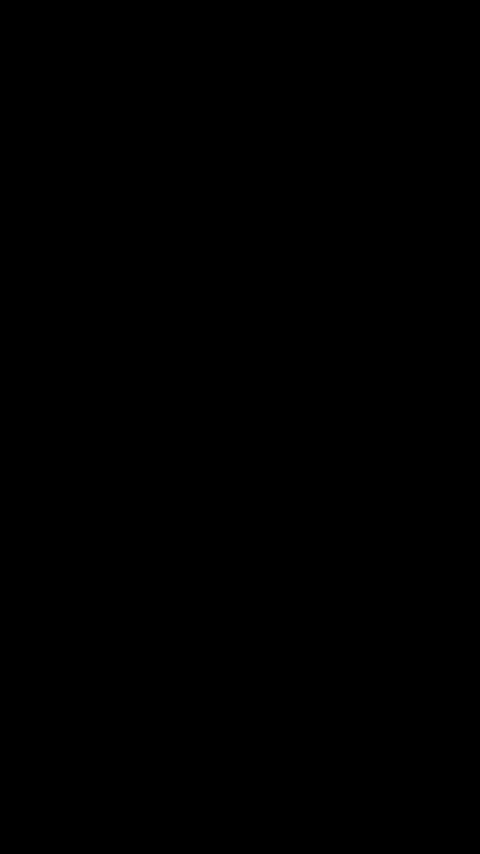}
\vfill 
}
\end{tcolorbox}

\begin{tcolorbox}[
colback=lightProxYellow!10,
colframe=lightProxYellow,
left=2mm, right=2mm, top=1mm, bottom=1mm,
before skip=0pt, after skip=1pt,
title=\textcolor{black}{turn9: read\_media\_file}]
{\small
"path": workspace/video\_frames/sec\_0010.jpg
}
\end{tcolorbox}

\begin{tcolorbox}[
colback=lightProxPink!10,
colframe=lightProxPink,
left=2mm, right=2mm, top=1mm, bottom=1mm,
before skip=0pt, after skip=1pt,
title=\textcolor{black}{Tool Response}]
{\small
\includegraphics[width=0.3\linewidth, keepaspectratio]{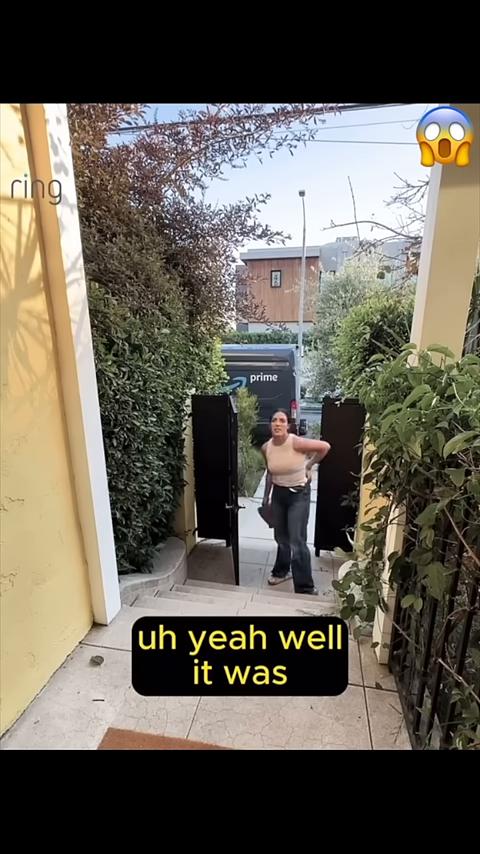}
\vfill 
}
\end{tcolorbox}

\begin{tcolorbox}[
colback=lightProxYellow!10,
colframe=lightProxYellow,
left=2mm, right=2mm, top=1mm, bottom=1mm,
before skip=0pt, after skip=1pt,
title=\textcolor{black}{turn10: read\_media\_file}]
{\small
"path": workspace/video\_frames/sec\_0020.jpg
}
\end{tcolorbox}

\begin{tcolorbox}[
colback=lightProxPink!10,
colframe=lightProxPink,
left=2mm, right=2mm, top=1mm, bottom=1mm,
before skip=0pt, after skip=1pt,
title=\textcolor{black}{Tool Response}]
{\small
\includegraphics[width=0.3\linewidth, keepaspectratio]{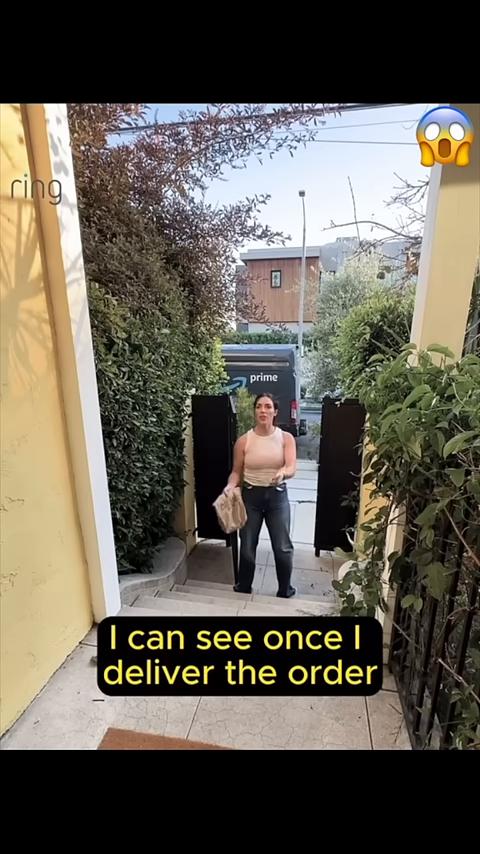}
\vfill 
}
\end{tcolorbox}

\begin{tcolorbox}[
colback=lightProxYellow!10,
colframe=lightProxYellow,
left=2mm, right=2mm, top=1mm, bottom=1mm,
before skip=0pt, after skip=1pt,
title=\textcolor{black}{turn11: read\_media\_file}]
{\small
"path": workspace/video\_frames/sec\_0030.jpg
}
\end{tcolorbox}

\begin{tcolorbox}[
colback=lightProxPink!10,
colframe=lightProxPink,
left=2mm, right=2mm, top=1mm, bottom=1mm,
before skip=0pt, after skip=1pt,
title=\textcolor{black}{Tool Response}]
{\small
\includegraphics[width=0.3\linewidth, keepaspectratio]{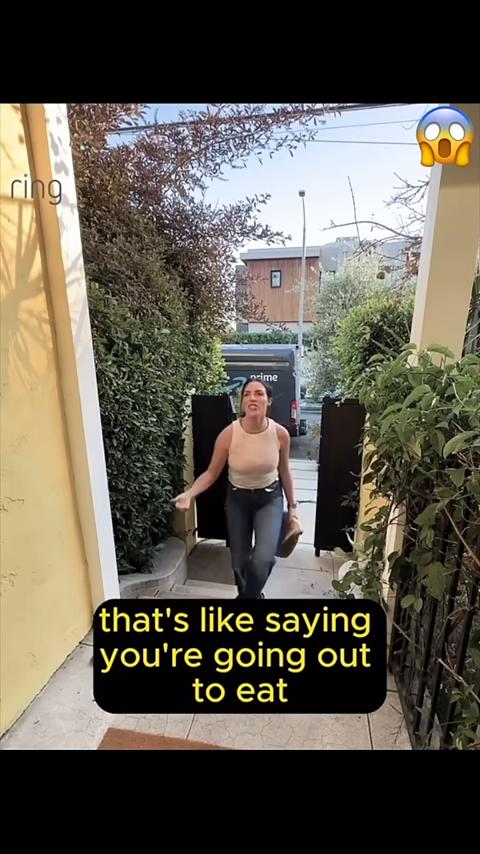}
\vfill 
}
\end{tcolorbox}

\begin{tcolorbox}[
colback=lightProxYellow!10,
colframe=lightProxYellow,
left=2mm, right=2mm, top=1mm, bottom=1mm,
before skip=0pt, after skip=1pt,
title=\textcolor{black}{turn12: read\_media\_file}]
{\small
"path": workspace/video\_frames/sec\_0040.jpg
}
\end{tcolorbox}

\begin{tcolorbox}[
colback=lightProxPink!10,
colframe=lightProxPink,
left=2mm, right=2mm, top=1mm, bottom=1mm,
before skip=0pt, after skip=1pt,
title=\textcolor{black}{Tool Response}]
{\small
\includegraphics[width=0.3\linewidth, keepaspectratio]{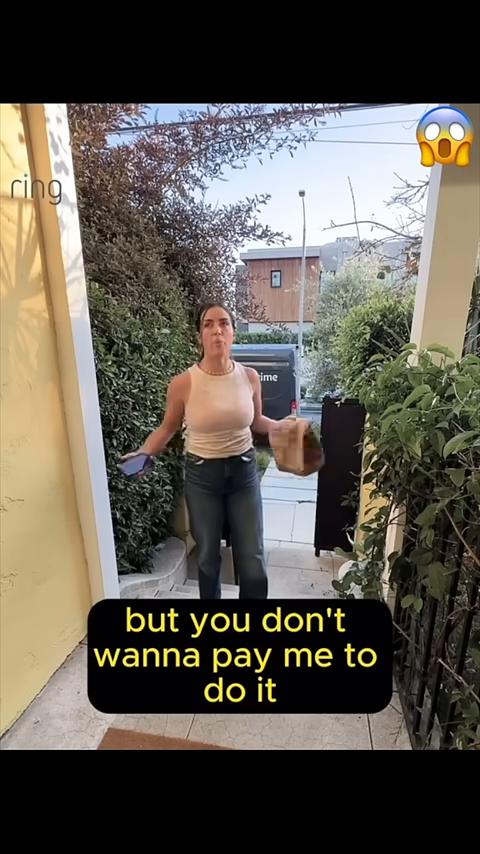}
\vfill 
}
\end{tcolorbox}

\begin{tcolorbox}[
colback=lightProxYellow!10,
colframe=lightProxYellow,
left=2mm, right=2mm, top=1mm, bottom=1mm,
before skip=0pt, after skip=1pt,
title=\textcolor{black}{turn13: read\_media\_file}]
{\small
"path": workspace/video\_frames/sec\_0050.jpg
}
\end{tcolorbox}

\begin{tcolorbox}[
colback=lightProxPink!10,
colframe=lightProxPink,
left=2mm, right=2mm, top=1mm, bottom=1mm,
before skip=0pt, after skip=1pt,
title=\textcolor{black}{Tool Response}]
{\small
\includegraphics[width=0.3\linewidth, keepaspectratio]{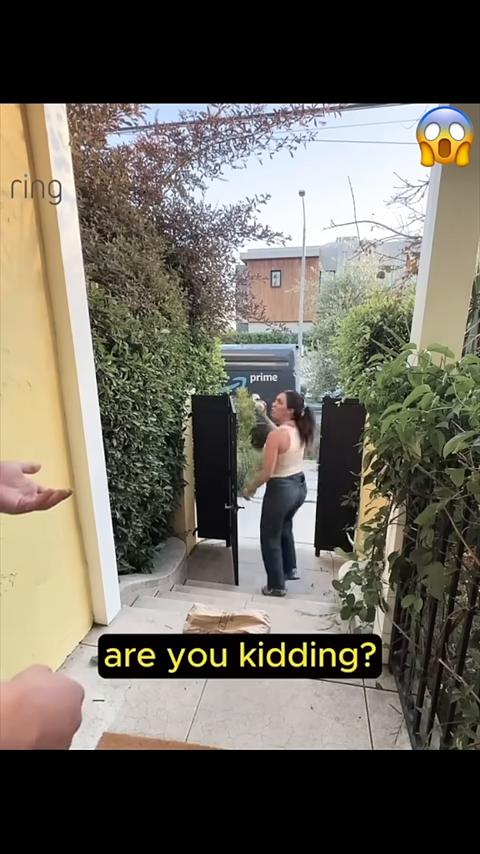}
\vfill 
}
\end{tcolorbox}

\begin{tcolorbox}[
colback=lightProxYellow!10,
colframe=lightProxYellow,
left=2mm, right=2mm, top=1mm, bottom=1mm,
before skip=0pt, after skip=1pt,
title=\textcolor{black}{turn14: read\_media\_file}]
{\small
"path": workspace/video\_frames/sec\_0055.jpg
}
\end{tcolorbox}

\begin{tcolorbox}[
colback=lightProxPink!10,
colframe=lightProxPink,
left=2mm, right=2mm, top=1mm, bottom=1mm,
before skip=0pt, after skip=1pt,
title=\textcolor{black}{Tool Response}]
{\small
\includegraphics[width=0.3\linewidth, keepaspectratio]{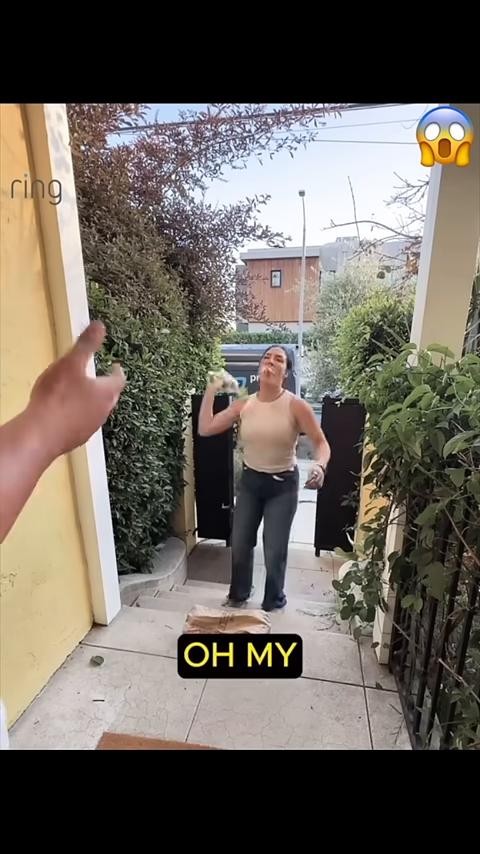}
\vfill 
}
\end{tcolorbox}

In the task Customer\_Service\_TAN/Food\_Delivery-delivery\_problems\_report, the agent needs to first verify the courier's identity by retrieving data from an Excel order table based on the provided order number. Subsequently, the agent is required to perform a frame-by-frame analysis of the video evidence to objectively record violations such as food damage or verbal conflicts, reconstruct the cause of the dispute (e.g., a tip-related confrontation), and ultimately generate a formal investigation report titled incident\_report.docx following a predefined structure that includes penalty and compensation recommendations. \\
During the execution process, due to a fixed-frame extraction strategy (every 10 seconds) and a lack of deep video causality modeling, the agent (MiniMax M2.5) failed to capture or understand the continuous actions of the courier within the temporal dimension. This led the agent to completely miss critical aggressive behaviors, such as the deliberate throwing of food and the confrontation over a tip, resulting in the false conclusion that there was no evidence of aggressive behavior.

\subsubsection{Cross-modal Information Fusion}
\begin{tcolorbox}[
colback=lightProxLightBlue!10,
colframe=lightProxLightBlue,
left=2mm, right=2mm, top=1mm, bottom=1mm,
before skip=0pt, after skip=1pt,
title=\textcolor{black}{Task Context}]

{\small
\textbf{User Request:}
\\I am a content creator who wants to share my real experience and explain why I do not recommend working as an UberEats courier on Instagram.
I will provide a selfie-style video in which I explain three reasons why I believe working for UberEats is not worth it. Please summarize and refine these three reasons, write an Instagram caption, and organize the written content into a Word document. When I discuss the first reason, there is a screenshot showing my working hours in the video — please capture that frame and save it as a separate image file.\\
\textbf{Domain Rules:}
\\1. Video Understanding: The agent must fully watch the input video and identify exactly three reasons stated by the speaker explaining why working for UberEats is not recommended. No additional reasons may be inferred or added.
\\2. Reason Condensation: Each reason must be summarized into one concise, impactful sentence suitable for social media consumption.
\\3. Image Extraction: When the first reason is being discussed, if the video displays any work-hour or income-related statistics interface, the agent must capture that frame and save it as a standalone image file named \texttt{UberEats\_WorkHours\_Screenshot.png}.
\\4. Final Output: The agent must generate a complete Instagram caption and save all written content into a Word document named \texttt{UberEats\_Instagram\_Content.docx}. The image file must not be embedded in the Word document.
}
\end{tcolorbox}

\begin{tcolorbox}[
colback=lightProxYellow!10,
colframe=lightProxYellow,
left=2mm, right=2mm, top=1mm, bottom=1mm,
before skip=0pt, after skip=1pt,
title=\textcolor{black}{turn4: audio\_to\_text}]
{\small
"path": workspace/UberEats\_audio.mp3
}
\end{tcolorbox}

\begin{tcolorbox}[
colback=lightProxPink!10,
colframe=lightProxPink,
left=2mm, right=2mm, top=1mm, bottom=1mm,
before skip=0pt, after skip=1pt,
title=\textcolor{black}{Tool Response}]
{\small
"srt\_text": "Here are the top three reasons not to work for food delivery services like door-dash and Uber eats. The first is that unlike a normal job, there's no overtime. I work a little over 60 hours a week right now and although I'm definitely making a lot more of my last job, I'm working way longer hours and getting paid the same no matter how much I work. The second reason is you're very dependent on tips. More than half of what I make is in the form of tips, so if you live an area ab... (truncated) 
}
\end{tcolorbox}

\begin{tcolorbox}[
colback=lightProxYellow!10,
colframe=lightProxYellow,
left=2mm, right=2mm, top=1mm, bottom=1mm,
before skip=0pt, after skip=1pt,
title=\textcolor{black}{turn6: export\_frames\_every\_second}]
{\small
"interval\_seconds":1,"output\_dir":"workspace/frames","video\_path":"workspace/UberEats.mp4"
}
\end{tcolorbox}

\begin{tcolorbox}[
colback=lightProxPink!10,
colframe=lightProxPink,
left=2mm, right=2mm, top=1mm, bottom=1mm,
before skip=0pt, after skip=1pt,
title=\textcolor{black}{Tool Response}]
{\small
"exported\_count":42,
"output\_directory":"workspace/frames",\\ 
"frames": ["workspace/frames/sec\_0000.jpg", "workspace/frames/sec\_0001.jpg",\\
"workspace/frames/sec\_0002.jpg", ... (truncated)
}
\end{tcolorbox}

\begin{tcolorbox}[
colback=lightProxYellow!10,
colframe=lightProxYellow,
left=2mm, right=2mm, top=1mm, bottom=1mm,
before skip=0pt, after skip=1pt,
title=\textcolor{black}{turn7: read\_media\_file}]
{\small
"path":"workspace/frames/sec\_0005.jpg"
}
\end{tcolorbox}

\begin{tcolorbox}[
colback=lightProxPink!10,
colframe=lightProxPink,
left=2mm, right=2mm, top=1mm, bottom=1mm,
before skip=0pt, after skip=1pt,
title=\textcolor{black}{Tool Response}]
{\small
\includegraphics[width=0.15\linewidth, keepaspectratio]{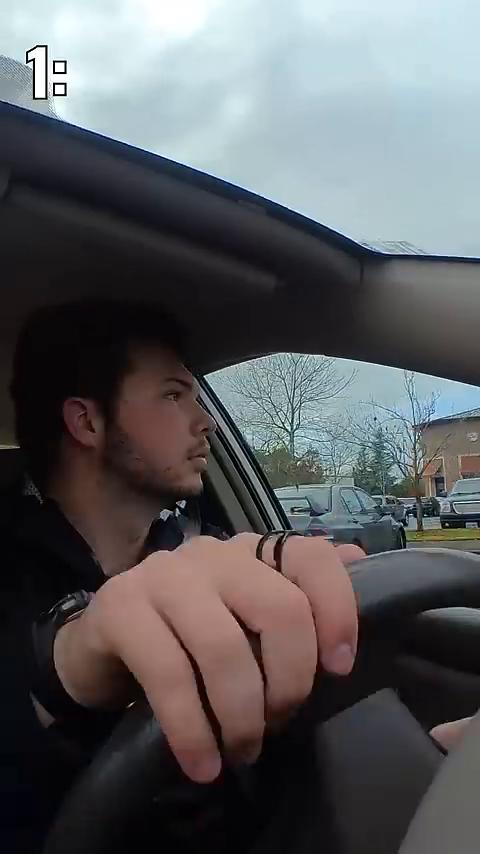}
\vfill 
}
\end{tcolorbox}

\begin{tcolorbox}[
colback=lightProxYellow!10,
colframe=lightProxYellow,
left=2mm, right=2mm, top=1mm, bottom=1mm,
before skip=0pt, after skip=1pt,
title=\textcolor{black}{turn11: read\_multiple\_files}]
{\small
"paths":["workspace/frames/sec\_0018.jpg","workspace/frames/sec\_0040.jpg"]
}
\end{tcolorbox}

\begin{tcolorbox}[
colback=lightProxPink!10,
colframe=lightProxPink,
left=2mm, right=2mm, top=1mm, bottom=1mm,
before skip=0pt, after skip=1pt,
title=\textcolor{black}{Tool Response}]
\includegraphics[width=0.3\linewidth, keepaspectratio]{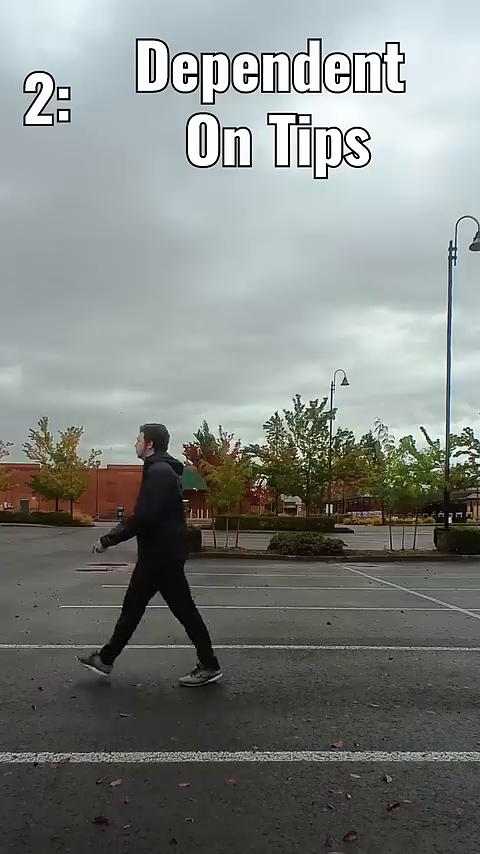}
\includegraphics[width=0.3\linewidth, keepaspectratio]{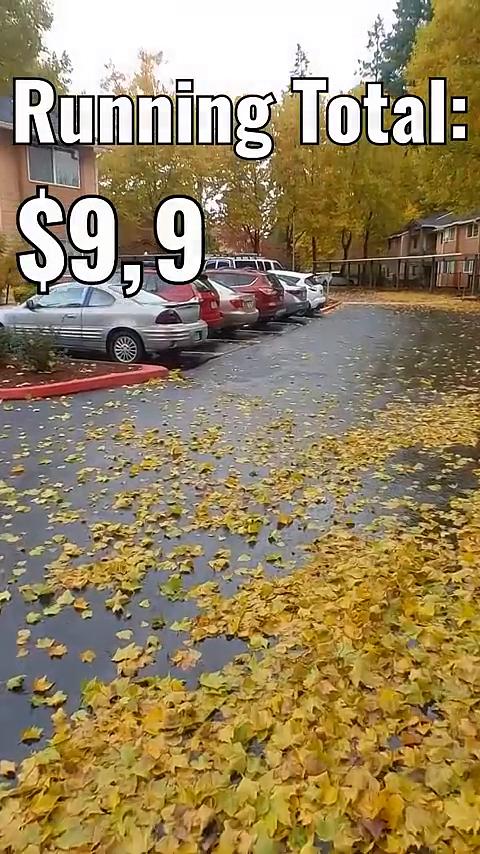}

\end{tcolorbox}

\begin{tcolorbox}[
colback=lightProxYellow!10,
colframe=lightProxYellow,
left=2mm, right=2mm, top=1mm, bottom=1mm,
before skip=0pt, after skip=1pt,
title=\textcolor{black}{turn13: move\_file}]
{\small
"destination":"workspace/UberEats\_WorkHours\_Screenshot.png",\\"source":"workspace/frames/sec\_0040.jpg"
}
\end{tcolorbox}

\begin{figure}[H]
    \centering
    \includegraphics[width=\linewidth, height=0.3\textheight, keepaspectratio]{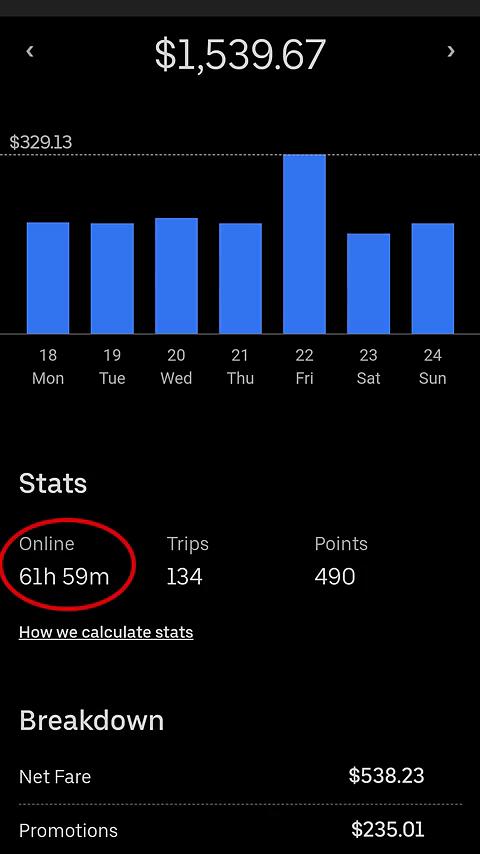}
    \caption{Ground truth screenshot displaying the work hours record interface.}
    \label{fig:wedding_layout_sketch}
\end{figure}

In the task \texttt{Customer\_Service/Food\_Delivery-ubereats\_negative}, the agent needs to act as a multimodal content assistant by processing a video clip, which first requires using speech recognition technology to accurately extract and condense the author's three core reasons for not recommending work as an UberEats courier. These reasons must be transformed into concise social media captions suitable for Instagram and organized into a Word document titled \texttt{UberEats\_Instagram\_Content.docx}. Simultaneously, the agent must demonstrate cross-modal alignment capabilities by identifying and capturing a data interface showing working hours or income statistics from the video during the specific segment where the author discusses the first reason, saving it as a standalone image file named \texttt{UberEats\_WorkHours\_Screenshot.png} while strictly adhering to naming conventions and file organization requirements.
\\During the execution process, Gemini 3.1 Pro successfully utilized an \texttt{audio\_to\_text} tool to obtain the transcript and identify the starting point of the first reason's discussion; however, it exhibited significant deficiencies in multimodal information fusion. Specifically, the agent failed to effectively map the semantic timeline of the audio to the visual timeline of the video, leading to a failure in the keyframe capture request. Instead of extracting the statistical interface from the beginning of the video, it erroneously captured a frame at the 40-second mark showing an unrelated outdoor scene. This failure in cross-modal alignment directly caused the task results to deviate severely from the user's requirements, exposing the model's insufficient depth in understanding complex audiovisual semantic correlations.

\clearpage
\subsubsection{Multimodal Hallucination}
\begin{tcolorbox}[
colback=lightProxLightBlue!10,
colframe=lightProxLightBlue,
left=2mm, right=2mm, top=1mm, bottom=1mm,
before skip=0pt, after skip=1pt,
title=\textcolor{black}{Task Context}]

{\small
\textbf{User Request:}
\\I am a QA tester preparing a high-signal bug report for a Minecraft duplication glitch that allows duplicating enchanted gear.
Use the provided gameplay recording to locate the first moment the item duplication glitch is clearly triggered. Extract three keyframes and write a standard Word bug report containing precise MM:SS timestamps and a step-by-step reproduction procedure. All generated files must comply with official QA file naming specifications.\\
\textbf{Domain Rules:}\\
Game QA Evidence Specialist\\
\\1. Glitch Moment Definition: The glitch moment refers to the earliest timestamp where duplicated enchanted items spontaneously appear in inventory or container interface without legitimate crafting, trading, or natural loot acquisition behavior.
\\2. Timestamp Identification: The agent must review the full video footage to confirm the exact first trigger time of the duplication glitch.
\\3. Keyframes: Extract exactly three PNG keyframes from the original video: (a) 2 seconds before the glitch trigger moment, (b) the exact glitch trigger moment, (c) 2 seconds after the glitch trigger moment. Save them as \texttt{glitch\_pre.png}, \texttt{glitch\_trigger.png}, \texttt{glitch\_post.png}.
\\4. Report: Create \texttt{mc\_item\_duplication\_bug\_report.docx} with fixed sections: Title, Test Environment (mark unknown items as \texttt{Not provided}), Numbered Reproduction Steps, Expected Behavior, Actual Behavior, and Evidence Attachment (list screenshot files with corresponding MM:SS timestamps). All timestamps must follow MM:SS standard format.
}
\end{tcolorbox}

\begin{tcolorbox}[
colback=lightProxYellow!10,
colframe=lightProxYellow,
left=2mm, right=2mm, top=1mm, bottom=1mm,
before skip=0pt, after skip=1pt,
title=\textcolor{black}{turn4: export\_frames\_every\_second}]
{\small
"video\_path":"workspace/bug1.mp4",
"interval\_seconds":1,
"output\_dir":"workspace/frames"
}
\end{tcolorbox}

\begin{tcolorbox}[
colback=lightProxPink!10,
colframe=lightProxPink,
left=2mm, right=2mm, top=1mm, bottom=1mm,
before skip=0pt, after skip=1pt,
title=\textcolor{black}{Tool Response}]
{\small
"exported\_count":9,\\"output\_directory": "workspace/frames",\\
"frames": [
"workspace/frames/sec\_0000.jpg",    "workspace/frames/sec\_0001.jpg",\\
"workspace/frames/sec\_0002.jpg",... (truncated)
}
\end{tcolorbox}

\begin{tcolorbox}[
colback=lightProxYellow!10,
colframe=lightProxYellow,
left=2mm, right=2mm, top=1mm, bottom=1mm,
before skip=0pt, after skip=1pt,
title=\textcolor{black}{turn5: read\_media\_file}]
{\small
"path":"workspace/frames/sec\_0000.jpg"
}
\end{tcolorbox}

\begin{tcolorbox}[
colback=lightProxPink!10,
colframe=lightProxPink,
left=2mm, right=2mm, top=1mm, bottom=1mm,
before skip=0pt, after skip=1pt,
title=\textcolor{black}{Tool Response}]
{\small
\includegraphics[width=0.25\linewidth, keepaspectratio]{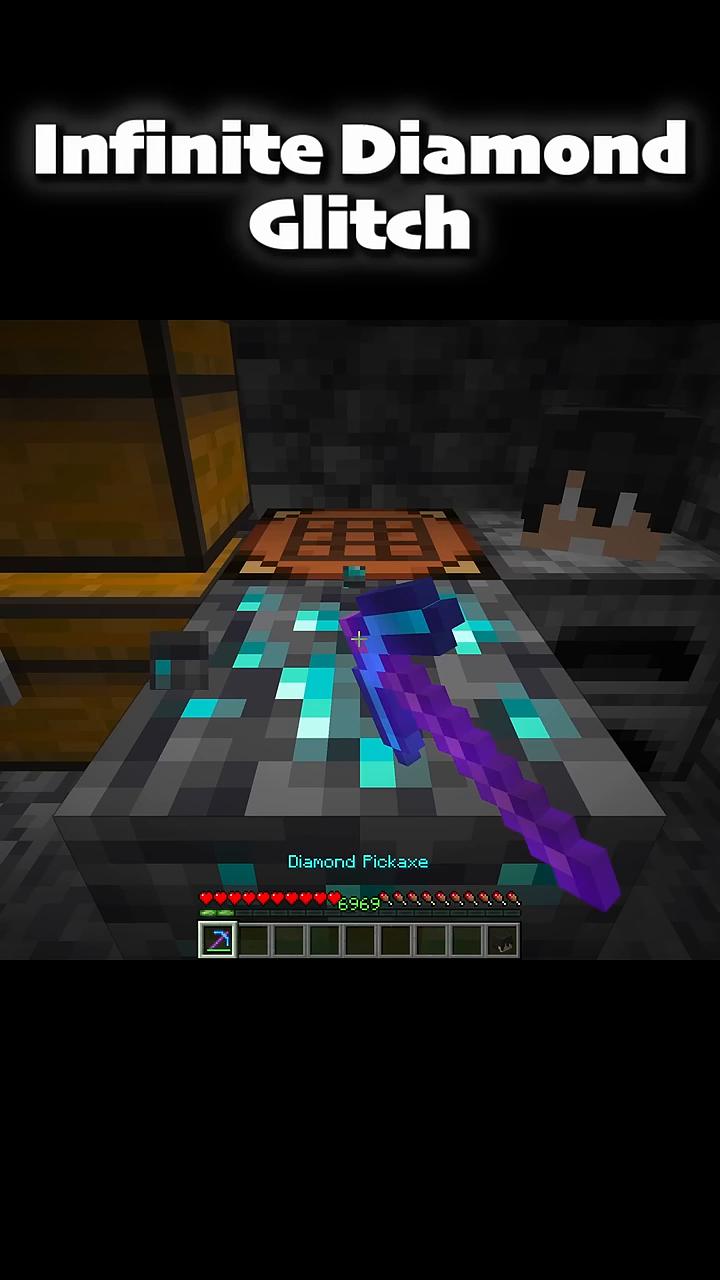}
\vfill 
}
\end{tcolorbox}

\begin{tcolorbox}[
colback=lightProxYellow!10,
colframe=lightProxYellow,
left=2mm, right=2mm, top=1mm, bottom=1mm,
before skip=0pt, after skip=1pt,
title=\textcolor{black}{turn6: export\_frame\_at\_second}]
{\small
"second":0,
"video\_path":"workspace/bug1.mp4",
"output\_path":"workspace/frame\_before.png"
}
\end{tcolorbox}

\begin{tcolorbox}[
colback=lightProxPink!10,
colframe=lightProxPink,
left=2mm, right=2mm, top=1mm, bottom=1mm,
before skip=0pt, after skip=1pt,
title=\textcolor{black}{Tool Response}]
{\small
workspace/frame\_before.png
}
\end{tcolorbox}

\begin{tcolorbox}[
colback=lightProxYellow!10,
colframe=lightProxYellow,
left=2mm, right=2mm, top=1mm, bottom=1mm,
before skip=0pt, after skip=1pt,
title=\textcolor{black}{turn7: create\_document}]
{\small
"filename":"workspace/mc\_infinite\_diamonds\_bug\_report.docx"
}
\end{tcolorbox}

In the task \texttt{Intelligent\_Creation/Game-bug\_report}, the agent needs to precisely locate the trigger moment of a Minecraft item duplication glitch by reviewing gameplay footage to identify the exact second enchanted gear abnormally appears. The agent must then extract three PNG keyframes representing the moments before, during, and after the trigger and generate a standardized report titled \texttt{mc\_item\_duplication\_bug\_report.docx} that includes precise MM:SS timestamps, detailed reproduction steps, and an evidence attachment list.\\
During the execution process, the GPT 4o exhibited severe hallucinations by skipping the actual visual reasoning process after viewing only the first frame of the video. Consequently, it fabricated the glitch trigger moment and all subsequent reproduction steps and timestamp data, failing to fulfill the domain rule of confirming the actual trigger time.

\subsection{COMPLETE EXAMPLE TASK TRAJECTORIES}
\label{app:trajectories}

\subsubsection{Traj 1}

\begin{tcolorbox}[
colback=lightProxLightBlue!10,
colframe=lightProxLightBlue,
left=2mm, right=2mm, after skip=1pt, title=\textcolor{black}{Task Context}]

{\small
\textbf{User Request:}
\\I am the General Manager of Nebula Heights Resort.
\\We have a very important guest checking in today: a boy named Leo who is celebrating his 10th birthday and loves astronauts and space exploration. First, I would like you to use the photo from Leo's parents, \texttt{Leo.png}, to generate an image of Leo wearing a spacesuit, standing on the moon, with a spaceship next to him and a blue starry sky as the background. This image should be named \texttt{Leo\_astronaut.png}. Then, to welcome Leo, I need you to create a \texttt{welcome\_tv.pptx} for the in-room TV. The PPT's \texttt{Welcome Title} should be \texttt{``Space Explorer''}, and the image you just created should be inserted into the position specified by the template. The \texttt{Greeting} section should say \texttt{``Happy 10th Birthday, Astronaut Leo!''}. Finally, please insert the provided \texttt{hotel\_logo.png} into the PPTX.

\textbf{Domain Rules:}
\\Creative Experience Designer
\\1. Visual Consistency: When designing the hotel's welcome-themed PPTX, you should first review the template file \texttt{example.pptx} and then design the PPTX according to the template's layout. After completing the PPT, you should check that the layout is consistent and the design is visually appealing, and adjust the PPT as needed. 
\\2. If the generated image does not meet the user's actual requirements, you need to re-generate the image until it meets the user's requirements.
\\3. PPTx details: The text for the \texttt{Welcome Title} should be set to \texttt{32 pt}. The \texttt{Greeting} section text should be set to \texttt{24 pt}. The provided \texttt{hotel\_logo.png} must be inserted at the bottom of the slide, and the image should not be overly eye-catching.
}
\end{tcolorbox}

\begin{tcolorbox}[
colback=lightProxGreen!10,
colframe=lightProxGreen,
left=2mm, right=2mm, top=1mm, bottom=1mm,
before skip=0pt, after skip=1pt,
title=\textcolor{black}{Thinking}]
{\small
Let me break down this task:

1. First, I need to generate an image of Leo wearing a spacesuit, standing on the moon, with a spaceship next to him and a blue starry sky as the background. The image should be named "Leo\_astronaut.png".

2. Then, I need to create a welcome PPTX ("welcome\_tv.pptx") using the template "example.pptx":
   - Welcome Title: "Space Explorer"
   - Insert the generated image ("Leo\_astronaut.png") into the position specified by the template
   - Greeting section: "Happy 10th Birthday, Astronaut Leo!"
   - Insert "hotel\_logo.png" at the bottom of the slide

Let me start by:
1. Reading the template file "example.pptx" to understand its layout
2. Reading the "Leo.png" and "hotel\_logo.png" files to understand what we're working with
3. Generating the astronaut image

Let me first check the workspace directory structure and read the necessary files.
}
\end{tcolorbox}

\begin{tcolorbox}[
colback=lightProxYellow!10,
colframe=lightProxYellow,
left=2mm, right=2mm, top=1mm, bottom=1mm,
before skip=0pt, after skip=1pt,
title=\textcolor{black}{list\_directory}]
{\small
"path": "workspace"
}
\end{tcolorbox}

\begin{tcolorbox}[
colback=lightProxPink!10,
colframe=lightProxPink,
left=2mm, right=2mm, top=1mm, bottom=1mm,
before skip=0pt, after skip=1pt,
title=\textcolor{black}{Tool Response}]
{\small
\textcolor{blue}{[FILE]} Leo.png\\ \textcolor{blue}{[FILE]} example.pptx\\ \textcolor{blue}{[FILE]} hotel\_logo.png
}
\end{tcolorbox}

\begin{tcolorbox}[
colback=lightProxYellow!10,
colframe=lightProxYellow,
left=2mm, right=2mm, top=1mm, bottom=1mm,
before skip=0pt, after skip=1pt,
title=\textcolor{black}{read\_media\_file}]
{\small
"path": "workspace/Leo.png"
}
\end{tcolorbox}

\begin{tcolorbox}[
colback=lightProxPink!10,
colframe=lightProxPink,
left=2mm, right=2mm, top=1mm, bottom=1mm,
before skip=0pt, after skip=1pt,
title=\textcolor{black}{Tool Response}]
{\small
\includegraphics[width=0.5\linewidth, keepaspectratio]{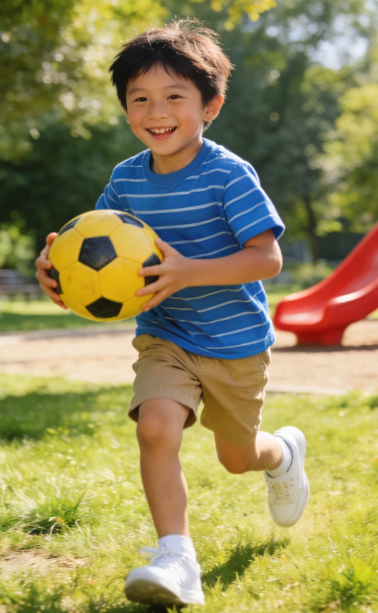}
\vfill 
\label{fig:ppt_hotel_leo}
}
\end{tcolorbox}

\begin{tcolorbox}[
colback=lightProxYellow!10,
colframe=lightProxYellow,
left=2mm, right=2mm, top=1mm, bottom=1mm,
before skip=0pt, after skip=1pt,
title=\textcolor{black}{read\_media\_file}]
{\small
"path": "workspace/hotel\_logo.png"
}
\end{tcolorbox}

\begin{tcolorbox}[
colback=lightProxPink!10,
colframe=lightProxPink,
left=2mm, right=2mm, top=1mm, bottom=1mm,
before skip=0pt, after skip=1pt,
title=\textcolor{black}{Tool Response}]
{\small
\includegraphics[width=0.5\linewidth, keepaspectratio]{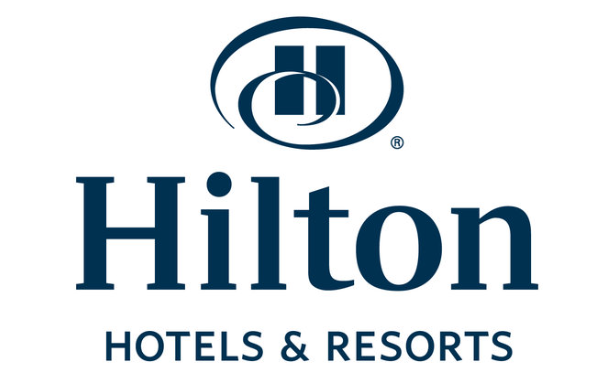}
\vfill 
\label{fig:ppt_hotel_logo}
}
\end{tcolorbox}

\begin{tcolorbox}[
colback=lightProxYellow!10,
colframe=lightProxYellow,
left=2mm, right=2mm, top=1mm, bottom=1mm,
before skip=0pt, after skip=1pt,
title=\textcolor{black}{get\_template\_file\_info}]
{\small
"template\_path": "workspace/example.pptx"
}
\end{tcolorbox}

\begin{tcolorbox}[
colback=lightProxPink!10,
colframe=lightProxPink,
left=2mm, right=2mm, top=1mm, bottom=1mm,
before skip=0pt, after skip=1pt,
title=\textcolor{black}{Tool Response}]
{\small
\{
  "template\_path": "workspace/example.pptx",
  "file\_size\_bytes": 34819,
  "slide\_count": 1,
  "layout\_count": 11,
  "slide\_layouts": [
    \{
      "index": 0,
      "name": "Title Slide",
      "placeholder\_count": 5
    \},
    \{
      "index": 1,
      "name": "Title and Content",
      "placeholder\_count": 5
    \},
    \{
      "index": 2,
      "name": "Section Header",
      "placeholder\_count": 5
    \},
    \{
      "index": 3,
      "name": "Two Content",... (truncated)
}
\end{tcolorbox}

\begin{tcolorbox}[
colback=lightProxGreen!10,
colframe=lightProxGreen,
left=2mm, right=2mm, top=1mm, bottom=1mm,
before skip=0pt, after skip=1pt,
title=\textcolor{black}{Thinking}]
{\small
Now I have a clear picture of the template and the images. Let me analyze:

1. **Leo.png**: Shows a young boy in a blue striped shirt holding a yellow soccer ball, running on grass with a playground in the background.

2. **hotel\_logo.png**: The Hilton Hotels \& Resorts logo.

3. **Template (example.pptx)**: Has 11 layouts. I need to understand the structure better. The template has layouts like "Title Slide", "Title and Content", etc.

Now let me:
1. First generate the astronaut image of Leo on the moon with a spaceship and blue starry sky
2. Then create the PPTX using the template

Let me start by generating the astronaut image.
}
\end{tcolorbox}

\begin{tcolorbox}[
colback=lightProxYellow!10,
colframe=lightProxYellow,
left=2mm, right=2mm, top=1mm, bottom=1mm,
before skip=0pt, after skip=1pt,
title=\textcolor{black}{image2image\_generate}]
{\small
"image\_aspect\_ratio": "16:9",
"image\_save\_path": "workspace/Leo\_astronaut.png",
"input\_image\_path": "workspace/Leo.png",
"prompt": "Transform this boy into an astronaut wearing a white spacesuit with helmet, standing on the moon's gray cratered surface. Next to him should be a futuristic silver spaceship. The background should be a beautiful blue starry sky with stars and possibly Earth visible in the distance. Keep the boy's face visible and recognizable, maintaining his cheerful expression. Create a magical space exploration scene suitable for a 10th birthday celebration."
}
\end{tcolorbox}

\begin{tcolorbox}[
colback=lightProxPink!10,
colframe=lightProxPink,
left=2mm, right=2mm, top=1mm, bottom=1mm,
before skip=0pt, after skip=1pt,
title=\textcolor{black}{Tool Response}]
{\small
\{
  "success": true,
  "message": "Image generated successfully.",
  "content": null,
  "output\_path": "workspace/Leo\_astronaut.png"
\}
}
\end{tcolorbox}

\begin{tcolorbox}[
colback=lightProxGreen!10,
colframe=lightProxGreen,
left=2mm, right=2mm, top=1mm, bottom=1mm,
before skip=0pt, after skip=1pt,
title=\textcolor{black}{Thinking}]
{\small
The astronaut image has been generated. Let me verify it looks good before proceeding with the PPTX creation.
}
\end{tcolorbox}

\begin{tcolorbox}[
colback=lightProxYellow!10,
colframe=lightProxYellow,
left=2mm, right=2mm, top=1mm, bottom=1mm,
before skip=0pt, after skip=1pt,
title=\textcolor{black}{read\_media\_file}]
{\small
"path": "workspace/Leo\_astronaut.png"
}
\end{tcolorbox}

\begin{tcolorbox}[
colback=lightProxPink!10,
colframe=lightProxPink,
left=2mm, right=2mm, top=1mm, bottom=1mm,
before skip=0pt, after skip=1pt,
title=\textcolor{black}{Tool Response}]
{\small
\includegraphics[width=0.5\linewidth, keepaspectratio]{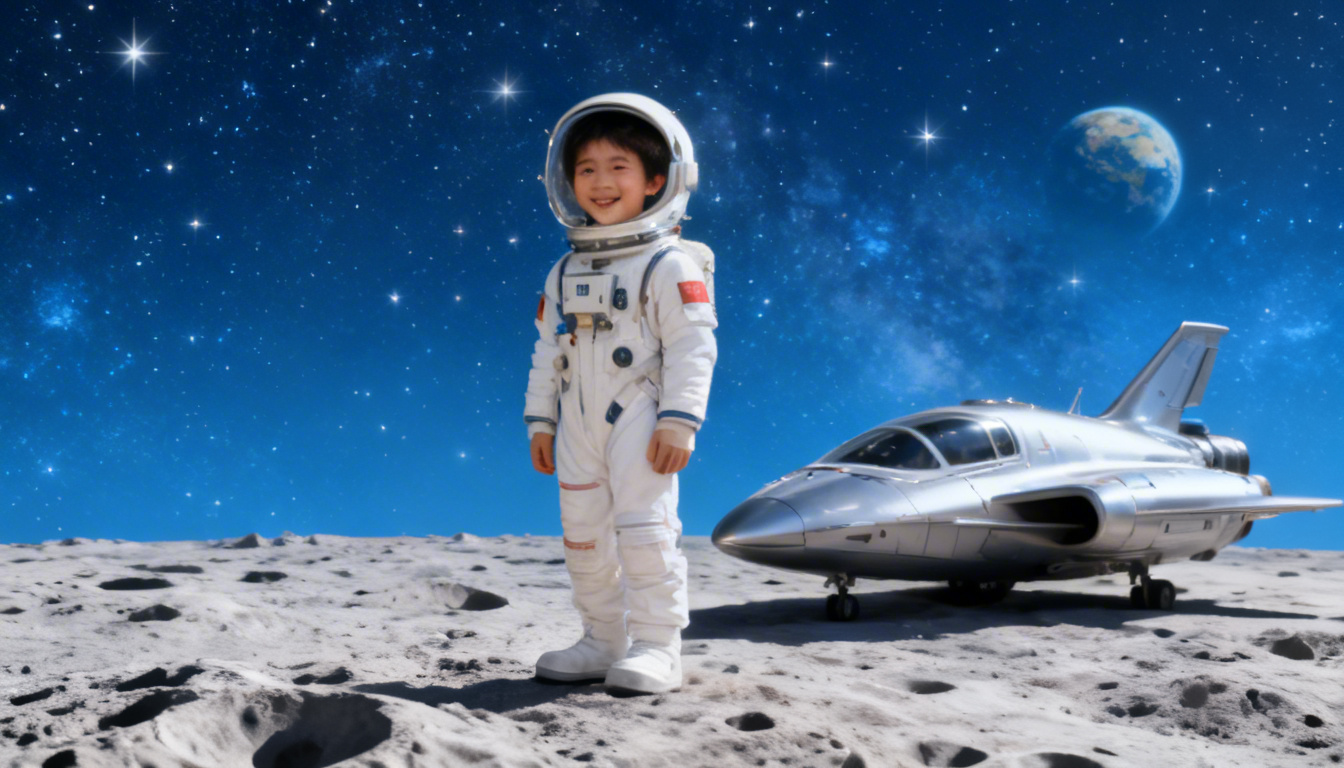}
\vfill
}
\end{tcolorbox}

\begin{tcolorbox}[
colback=lightProxGreen!10,
colframe=lightProxGreen,
left=2mm, right=2mm, top=1mm, bottom=1mm,
before skip=0pt, after skip=1pt,
title=\textcolor{black}{Thinking}]
{\small
The astronaut image looks great - Leo is wearing a spacesuit, standing on the moon with a spaceship next to him and a blue starry sky background. Now I need to create the welcome PPTX using the template. Let me open the template and create the presentation.

Based on the template info, I'll use layout index 1 (Title and Content) which has placeholders for title and content. I need to:
1. Create a presentation from the template
2. Set the "Welcome Title" to "Space Explorer" (32 pt)
3. Add the greeting "Happy 10th Birthday, Astronaut Leo!" (24 pt)
4. Insert the Leo\_astronaut.png image
5. Insert the hotel\_logo.png at the bottom

Let me open the template and start building the presentation.
}
\end{tcolorbox}

\begin{tcolorbox}[
colback=lightProxYellow!10,
colframe=lightProxYellow,
left=2mm, right=2mm, top=1mm, bottom=1mm,
before skip=0pt, after skip=1pt,
title=\textcolor{black}{create\_presentation\_from\_template}]
{\small
"id": "welcome\_presentation",
"template\_path": "workspace/example.pptx"
}
\end{tcolorbox}

\begin{tcolorbox}[
colback=lightProxPink!10,
colframe=lightProxPink,
left=2mm, right=2mm, top=1mm, bottom=1mm,
before skip=0pt, after skip=1pt,
title=\textcolor{black}{Tool Response}]
{\small
\{
  "presentation\_id": "welcome\_presentation",
  "message": "Created new presentation from template 'workspace/example.pptx' with ID: welcome\_presentation",
  "template\_path": "workspace/example.pptx",
  "slide\_count": 1,
  "layout\_count": 11
\}
}
\end{tcolorbox}

\begin{tcolorbox}[
colback=lightProxGreen!10,
colframe=lightProxGreen,
left=2mm, right=2mm, top=1mm, bottom=1mm,
before skip=0pt, after skip=1pt,
title=\textcolor{black}{Thinking}]
{\small
Good, the presentation is created. Now let me get the slide info to understand the placeholders better, and then add the content.
}
\end{tcolorbox}

\begin{tcolorbox}[
colback=lightProxYellow!10,
colframe=lightProxYellow,
left=2mm, right=2mm, top=1mm, bottom=1mm,
before skip=0pt, after skip=1pt,
title=\textcolor{black}{get\_slide\_info}]
{\small
"presentation\_id": "welcome\_presentation",
"slide\_index": 0
}
\end{tcolorbox}

\begin{tcolorbox}[
colback=lightProxPink!10,
colframe=lightProxPink,
left=2mm, right=2mm, top=1mm, bottom=1mm,
before skip=0pt, after skip=1pt,
title=\textcolor{black}{Tool Response}]
{\small
\{
  "slide\_index": 0,
  "layout\_name": "Title Slide",
  "placeholder\_count": 1,
  "placeholders": [
    \{
      "idx": 1,
      "type": "SUBTITLE (4)",
      "name": "Subtitle 2"
    \}
  ],
  "shape\_count": 4,
  "shapes": [
    \{
      "index": 0,
      "name": "Subtitle 2",
      "shape\_type": "PLACEHOLDER (14)",
      "left": 3742267,
      "top": 664105,
      "width": 3953933,
      "height": 563562
    \},... (truncated)
}
\end{tcolorbox}

\begin{tcolorbox}[
colback=lightProxGreen!10,
colframe=lightProxGreen,
left=2mm, right=2mm, top=1mm, bottom=1mm,
before skip=0pt, after skip=1pt,
title=\textcolor{black}{Thinking}]
{\small
I see the slide has placeholders. Let me add the title and greeting text. I'll use the manage\_text tool to add the "Space Explorer" title and the greeting. I also need to add the astronaut image and the hotel logo.

Let me first add the title text, then the greeting, then the images.
}
\end{tcolorbox}

\begin{tcolorbox}[
colback=lightProxYellow!10,
colframe=lightProxYellow,
left=2mm, right=2mm, top=1mm, bottom=1mm,
before skip=0pt, after skip=1pt,
title=\textcolor{black}{manage\_text}]
{\small
"alignment": "center",
"font\_size": "32",
"operation": "add",
"presentation\_id": "welcome\_presentation",
"slide\_index": 0,
"text": "Space Explorer",
"top": 0.5,
"width": 9
}
\end{tcolorbox}

\begin{tcolorbox}[
colback=lightProxPink!10,
colframe=lightProxPink,
left=2mm, right=2mm, top=1mm, bottom=1mm,
before skip=0pt, after skip=1pt,
title=\textcolor{black}{Tool Response}]
{\small
\{
  "message": "Added text box to slide 0",
  "shape\_index": 4,
  "text": "Space Explorer"
\}
}
\end{tcolorbox}

\begin{tcolorbox}[
colback=lightProxGreen!10,
colframe=lightProxGreen,
left=2mm, right=2mm, top=1mm, bottom=1mm,
before skip=0pt, after skip=1pt,
title=\textcolor{black}{Thinking}]
{\small
Good, the title is added. Now let me add the greeting text below it.
}
\end{tcolorbox}

\begin{tcolorbox}[
colback=lightProxYellow!10,
colframe=lightProxYellow,
left=2mm, right=2mm, top=1mm, bottom=1mm,
before skip=0pt, after skip=1pt,
title=\textcolor{black}{manage\_text}]
{\small
"alignment": "center",
"font\_size": "24",
"operation": "add",
"presentation\_id": "welcome\_presentation",
"slide\_index": 0,
"text": "Happy 10th Birthday, Astronaut Leo!",
"top": 2,
"width": 9
}
\end{tcolorbox}

\begin{tcolorbox}[
colback=lightProxPink!10,
colframe=lightProxPink,
left=2mm, right=2mm, top=1mm, bottom=1mm,
before skip=0pt, after skip=1pt,
title=\textcolor{black}{Tool Response}]
{\small
\{
  "message": "Added text box to slide 0",
  "shape\_index": 5,
  "text": "Happy 10th Birthday, Astronaut Leo!"
\}
}
\end{tcolorbox}

\begin{tcolorbox}[
colback=lightProxGreen!10,
colframe=lightProxGreen,
left=2mm, right=2mm, top=1mm, bottom=1mm,
before skip=0pt, after skip=1pt,
title=\textcolor{black}{Thinking}]
{\small
Good, now I need to add the astronaut image and the hotel logo. Let me add the Leo\_astronaut.png image first, then the hotel\_logo.png at the bottom.
}
\end{tcolorbox}

\begin{tcolorbox}[
colback=lightProxYellow!10,
colframe=lightProxYellow,
left=2mm, right=2mm, top=1mm, bottom=1mm,
before skip=0pt, after skip=1pt,
title=\textcolor{black}{manage\_image}]
{\small
"height": "4",
"image\_source": "workspace/Leo\_astronaut.png",
"left": 2.5,
"operation": "add",
"presentation\_id": "welcome\_presentation",
"slide\_index": 0,
"top": 3,
"width": "6"
}
\end{tcolorbox}

\begin{tcolorbox}[
colback=lightProxPink!10,
colframe=lightProxPink,
left=2mm, right=2mm, top=1mm, bottom=1mm,
before skip=0pt, after skip=1pt,
title=\textcolor{black}{Tool Response}]
{\small
\{
  "message": "Added image to slide 0",
  "shape\_index": 6,
  "image\_path": "workspace/Leo\_astronaut.png"
\}
}
\end{tcolorbox}

\begin{tcolorbox}[
colback=lightProxGreen!10,
colframe=lightProxGreen,
left=2mm, right=2mm, top=1mm, bottom=1mm,
before skip=0pt, after skip=1pt,
title=\textcolor{black}{Thinking}]
{\small
Good, the astronaut image is added. Now I need to add the hotel logo at the bottom of the slide. The logo should not be overly eye-catching, so I'll make it smaller.
}
\end{tcolorbox}

\begin{tcolorbox}[
colback=lightProxYellow!10,
colframe=lightProxYellow,
left=2mm, right=2mm, top=1mm, bottom=1mm,
before skip=0pt, after skip=1pt,
title=\textcolor{black}{manage\_image}]
{\small
"height": "0.8",
"image\_source": "workspace/hotel\_logo.png",
"left": 4.5,
"operation": "add",
"presentation\_id": "welcome\_presentation",
"slide\_index": 0,
"top": 7.5,
"width": "2"
}
\end{tcolorbox}

\begin{tcolorbox}[
colback=lightProxPink!10,
colframe=lightProxPink,
left=2mm, right=2mm, top=1mm, bottom=1mm,
before skip=0pt, after skip=1pt,
title=\textcolor{black}{Tool Response}]
{\small
\{
  "message": "Added image to slide 0",
  "shape\_index": 7,
  "image\_path": "workspace/hotel\_logo.png"
\}
}
\end{tcolorbox}

\begin{tcolorbox}[
colback=lightProxGreen!10,
colframe=lightProxGreen,
left=2mm, right=2mm, top=1mm, bottom=1mm,
before skip=0pt, after skip=1pt,
title=\textcolor{black}{Thinking}]
{\small
Good, now let me save the presentation to the specified file path.
}
\end{tcolorbox}

\begin{tcolorbox}[
colback=lightProxYellow!10,
colframe=lightProxYellow,
left=2mm, right=2mm, top=1mm, bottom=1mm,
before skip=0pt, after skip=1pt,
title=\textcolor{black}{save\_presentation}]
{\small
"file\_path": "workspace/welcome\_tv.pptx",
"presentation\_id": "welcome\_presentation"
}
\end{tcolorbox}

\begin{tcolorbox}[
colback=lightProxPink!10,
colframe=lightProxPink,
left=2mm, right=2mm, top=1mm, bottom=1mm,
before skip=0pt, after skip=1pt,
title=\textcolor{black}{Tool Response}]
{\small
\{
  "message": "Presentation saved to workspace/welcome\_tv.pptx",
  "file\_path": "workspace/welcome\_tv.pptx"
\}
}
\end{tcolorbox}

\begin{tcolorbox}[
colback=lightProxGreen!10,
colframe=lightProxGreen,
left=2mm, right=2mm, top=1mm, bottom=1mm,
before skip=0pt, after skip=1pt,
title=\textcolor{black}{Thinking}]
{\small
The presentation is saved. Let me verify the final result by extracting the text and checking the slide info to make sure everything is in place.
}
\end{tcolorbox}

\begin{tcolorbox}[
colback=lightProxYellow!10,
colframe=lightProxYellow,
left=2mm, right=2mm, top=1mm, bottom=1mm,
before skip=0pt, after skip=1pt,
title=\textcolor{black}{extract\_presentation\_text}]
{\small
"include\_slide\_info": true,
"presentation\_id": "welcome\_presentation"
}
\end{tcolorbox}

\begin{tcolorbox}[
colback=lightProxPink!10,
colframe=lightProxPink,
left=2mm, right=2mm, top=1mm, bottom=1mm,
before skip=0pt, after skip=1pt,
title=\textcolor{black}{Tool Response}]
{\small
\{
  "success": true,
  "presentation\_id": "welcome\_presentation",
  "total\_slides": 1,
  "slides\_with\_text": 1,
  "total\_text\_shapes": 1,
  "slides\_text": [
    \{
      "slide\_index": 0,
      "text\_content": \{
        "slide\_title": "",
        "placeholders": [
          \{
            "shape\_index": 0,
            "shape\_name": "Subtitle 2",
            "shape\_type": "PLACEHOLDER (14)",
            "text": "Welcome Title"
          \}
        ]
      \}
    \}
  ]
\}
}
\end{tcolorbox}

\begin{tcolorbox}[
colback=lightProxYellow!10,
colframe=lightProxYellow,
left=2mm, right=2mm, top=1mm, bottom=1mm,
before skip=0pt, after skip=1pt,
title=\textcolor{black}{get\_slide\_info}]
{\small
"presentation\_id": "welcome\_presentation",
"slide\_index": 0
}
\end{tcolorbox}

\begin{tcolorbox}[
colback=lightProxPink!10,
colframe=lightProxPink,
left=2mm, right=2mm, top=1mm, bottom=1mm,
before skip=0pt, after skip=1pt,
title=\textcolor{black}{Tool Response}]
{\small
\{
  "slide\_index": 0,
  "layout\_name": "Title Slide",
  "placeholder\_count": 1,
  "placeholders": [
    \{
      "idx": 1,
      "type": "SUBTITLE (4)",
      "name": "Subtitle 2"
    \}
  ],
  "shape\_count": 8,
  "shapes": [
    \{
      "index": 0,
      "name": "Subtitle 2",
      "shape\_type": "PLACEHOLDER (14)",
      "left": 3742267,
      "top": 664105,
      "width": 3953933,
      "height": 563562
    \},... (truncated)
}
\end{tcolorbox}

\begin{tcolorbox}[
colback=lightProxGreen!10,
colframe=lightProxGreen,
left=2mm, right=2mm, top=1mm, bottom=1mm,
before skip=0pt, after skip=1pt,
title=\textcolor{black}{Thinking}]
{\small
The presentation looks good! Let me verify the files are all in place by listing the directory:
}
\end{tcolorbox}

\begin{tcolorbox}[
colback=lightProxYellow!10,
colframe=lightProxYellow,
left=2mm, right=2mm, top=1mm, bottom=1mm,
before skip=0pt, after skip=1pt,
title=\textcolor{black}{list\_directory}]
{\small
"path": "workspace"
}
\end{tcolorbox}

\begin{tcolorbox}[
colback=lightProxPink!10,
colframe=lightProxPink,
left=2mm, right=2mm, top=1mm, bottom=1mm,
before skip=0pt, after skip=1pt,
title=\textcolor{black}{Tool Response}]
{\small
\textcolor{blue}{[FILE]} Leo.png\\ \textcolor{blue}{[FILE]} Leo\_astronaut.png\\ \textcolor{blue}{[FILE]} example.pptx\\ \textcolor{blue}{[FILE]} hotel\_logo.png\\ \textcolor{blue}{[FILE]} welcome\_tv.pptx
}
\end{tcolorbox}

\begin{tcolorbox}[
colback=lightProxGreen!10,
colframe=lightProxGreen,
left=2mm, right=2mm, top=1mm, bottom=1mm,
before skip=0pt, after skip=1pt,
title=\textcolor{black}{Thinking}]
{\small
All files are in place. Let me summarize what was accomplished:

1. Generated "Leo\_astronaut.png" - An image of Leo wearing a spacesuit, standing on the moon, with a spaceship next to him and a blue starry sky background.
2. Created "welcome\_tv.pptx" from the template.
3. Added "Space Explorer" as the Welcome Title (32 pt font).
4. Added "Happy 10th Birthday, Astronaut Leo!" as the Greeting (24 pt font).
5. Inserted the "Leo\_astronaut.png" image into the slide.
6. Inserted the "hotel\_logo.png" at the bottom of the slide (smaller size so it's not overly eye-catching).

The task is complete!
}
\end{tcolorbox}

\begin{tcolorbox}[
colback=lightProxYellow!10,
colframe=lightProxYellow,
left=2mm, right=2mm, top=1mm, bottom=1mm,
before skip=0pt, after skip=1pt,
title=\textcolor{black}{local-claim\_done}]
{\small
}
\end{tcolorbox}

\begin{tcolorbox}[
colback=lightProxPink!10,
colframe=lightProxPink,
left=2mm, right=2mm, top=1mm, bottom=1mm,
before skip=0pt, after skip=1pt,
title=\textcolor{black}{Tool Response}]
{\small
you have claimed the task is done!
}
\end{tcolorbox}

\subsubsection{Traj 2}
\begin{tcolorbox}[
colback=lightProxLightBlue!10,
colframe=lightProxLightBlue,
left=2mm, right=2mm, after skip=1pt, title=\textcolor{black}{Task Context}]
{\small
\textbf{User Request:}
\\I'm a software instructor at \texttt{Creative Tech Academy}.
Hello, I have recorded a short video (\texttt{Notebook.mp4}) showing how to use the \texttt{'pandas'} package to read an Excel table. Could you please listen to my video, and then record the two main parts of the tutorial: \texttt{Data Preparation} and \texttt{Code Example} in \texttt{tutorial.md} file? Each part should be started with a second-level header, and describe the content of each step in clear detail. Also, it's important to find the key supporting frame related to each part (\texttt{data\_preparation.png} for \texttt{Data Preparation}, \texttt{code\_example.png} for \texttt{Code Example}). Please save the captured images in the \texttt{assets} folder and insert them into the markdown document? The image titles are not needed.

\textbf{Domain Rules:}
\\Technical Writing Assistant
\\1. After taking the screenshot, please check that the screenshot content is complete and correct.
\\2. Please use Markdown syntax for inserting images and writing their captions.
}
\end{tcolorbox}

\begin{tcolorbox}[
colback=lightProxGreen!10,
colframe=lightProxGreen,
left=2mm, right=2mm, top=1mm, bottom=1mm,
before skip=0pt, after skip=1pt,
title=\textcolor{black}{Thinking}]
{\small
The user wants me to:\\
1. Listen to a video (`Notebook.mp4`) about using pandas to read an Excel table\\
2. Identify two main parts: `Data Preparation` and `Code Example`\\
3. Create a `tutorial.md` file with second-level headers for each part\\
4. Find key supporting frames:\\
   - `data\_preparation.png` for Data Preparation\\
   - `code\_example.png` for Code Example\\
5. Save the images in the `assets` folder\\
6. Insert the images into the markdown document\\
\\
Let me start by exploring the workspace to find the video file.
}
\end{tcolorbox}

\begin{tcolorbox}[
colback=lightProxYellow!10,
colframe=lightProxYellow,
left=2mm, right=2mm, top=1mm, bottom=1mm,
before skip=0pt, after skip=1pt,
title=\textcolor{black}{list\_directory}]
{\small
\{"path": "workspace/"\}
}
\end{tcolorbox}

\begin{tcolorbox}[
colback=lightProxPink!10,
colframe=lightProxPink,
left=2mm, right=2mm, top=1mm, bottom=1mm,
before skip=0pt, after skip=1pt,
title=\textcolor{black}{Tool Response}]
{\small
\textcolor{blue}{[FILE]} Notebook.mp4
}
\end{tcolorbox}

\begin{tcolorbox}[
colback=lightProxGreen!10,
colframe=lightProxGreen,
left=2mm, right=2mm, top=1mm, bottom=1mm,
before skip=0pt, after skip=1pt,
title=\textcolor{black}{Thinking}]
{\small
Found the video file. Let me first get the video metadata to understand its duration, and then I'll extract audio to transcribe it.
}
\end{tcolorbox}

\begin{tcolorbox}[
colback=lightProxYellow!10,
colframe=lightProxYellow,
left=2mm, right=2mm, top=1mm, bottom=1mm,
before skip=0pt, after skip=1pt,
title=\textcolor{black}{get\_video\_metadata\_json}]
{\small
\{
  "video\_path": "Notebook.mp4"
\}
}
\end{tcolorbox}

\begin{tcolorbox}[
colback=lightProxPink!10,
colframe=lightProxPink,
left=2mm, right=2mm, top=1mm, bottom=1mm,
before skip=0pt, after skip=1pt,
title=\textcolor{black}{Tool Response}]
{\small
\{\\
  "duration\_seconds": 46.52,\\
  "frame\_size": [\\
    1280,\\
    720\\
  ],\\
  "fps": 30.0,\\
  "approx\_total\_frames": 1396\\
\}
}
\end{tcolorbox}

\begin{tcolorbox}[
colback=lightProxGreen!10,
colframe=lightProxGreen,
left=2mm, right=2mm, top=1mm, bottom=1mm,
before skip=0pt, after skip=1pt,
title=\textcolor{black}{Thinking}]
{\small
The video is about 46 seconds long at 30 fps. Now let me extract the audio and transcribe it.
}
\end{tcolorbox}

\begin{tcolorbox}[
colback=lightProxYellow!10,
colframe=lightProxYellow,
left=2mm, right=2mm, top=1mm, bottom=1mm,
before skip=0pt, after skip=1pt,
title=\textcolor{black}{extract\_audio\_from\_video}]
{\small
\{\\
  "audio\_codec": "mp3",\\
  "output\_audio\_path": "notebook\_audio.mp3",\\
  "video\_path": "Notebook.mp4"\\
\}
}
\end{tcolorbox}

\begin{tcolorbox}[
colback=lightProxPink!10,
colframe=lightProxPink,
left=2mm, right=2mm, top=1mm, bottom=1mm,
before skip=0pt, after skip=1pt,
title=\textcolor{black}{Tool Response}]
{\small
Audio extracted successfully to notebook\_audio.mp3
}
\end{tcolorbox}

\begin{tcolorbox}[
colback=lightProxGreen!10,
colframe=lightProxGreen,
left=2mm, right=2mm, top=1mm, bottom=1mm,
before skip=0pt, after skip=1pt,
title=\textcolor{black}{Thinking}]
{\small
Now let me transcribe the audio to understand the content.
}
\end{tcolorbox}

\begin{tcolorbox}[
colback=lightProxYellow!10,
colframe=lightProxYellow,
left=2mm, right=2mm, top=1mm, bottom=1mm,
before skip=0pt, after skip=1pt,
title=\textcolor{black}{audio\_to\_text}]
{\small
\{\\
  "audio\_path": "notebook\_audio.mp3"\\
\}
}
\end{tcolorbox}

\begin{tcolorbox}[
colback=lightProxPink!10,
colframe=lightProxPink,
left=2mm, right=2mm, top=1mm, bottom=1mm,
before skip=0pt, after skip=1pt,
title=\textcolor{black}{Tool Response}]
{\small
\{\\
  "srt\_text": " Today, I'm going to show you how to import data contained in a CSV file into a Jupyter notebook. I have here CSV file that I've put into the same folder as my Jupyter notebook, as you can see. Now, I'm just going to go into my notebook and type Import Pandas. Then Pandas.Read under Score CSV, and then in parentheses, and quotes, the name of your CSV file. Now, I'm going to set a variable equivalent to that. Let's call our variable Data Frame. Now, we're going to ask Jupyter not... (truncated)"\\
\}
}
\end{tcolorbox}

\begin{tcolorbox}[
colback=lightProxGreen!10,
colframe=lightProxGreen,
left=2mm, right=2mm, top=1mm, bottom=1mm,
before skip=0pt, after skip=1pt,
title=\textcolor{black}{Thinking}]
{\small
Interesting! The audio actually talks about CSV files, not Excel files. But the user mentioned it's about reading an Excel table. Let me watch the video to see what's actually happening on screen - perhaps there's a mistake in the transcription or the user described it differently. Let me export some frames to understand what's actually being shown.\\
\\
Let me export frames at different points in the video to understand the content better.
}
\end{tcolorbox}

\begin{tcolorbox}[
colback=lightProxYellow!10,
colframe=lightProxYellow,
left=2mm, right=2mm, top=1mm, bottom=1mm,
before skip=0pt, after skip=1pt,
title=\textcolor{black}{export\_frames\_every\_second}]
{\small
\{\\
  "interval\_seconds": 5,\\
  "output\_dir": "frames",\\
  "video\_path": "Notebook.mp4"\\
\}
}
\end{tcolorbox}

\begin{tcolorbox}[
colback=lightProxPink!10,
colframe=lightProxPink,
left=2mm, right=2mm, top=1mm, bottom=1mm,
before skip=0pt, after skip=1pt,
title=\textcolor{black}{Tool Response}]
{\small
\{\\
  "exported\_count": 10,\\
  "output\_directory": "frames",\\
  "frames": [\\
    "frames/sec\_0000.jpg",\\
    "frames/sec\_0005.jpg",\\
    "frames/sec\_0010.jpg",\\
    "frames/sec\_... (truncated)"\\
\}
}
\end{tcolorbox}

\begin{tcolorbox}[
colback=lightProxGreen!10,
colframe=lightProxGreen,
left=2mm, right=2mm, top=1mm, bottom=1mm,
before skip=0pt, after skip=1pt,
title=\textcolor{black}{Thinking}]
{\small
Let me check the frames to see what's shown in the video. I'll look at a few key frames to understand the content better.
}
\end{tcolorbox}

\begin{tcolorbox}[
colback=lightProxYellow!10,
colframe=lightProxYellow,
left=2mm, right=2mm, top=1mm, bottom=1mm,
before skip=0pt, after skip=1pt,
title=\textcolor{black}{read\_media\_file}]
{\small
\{\\
  "path": "frames/sec\_0000.jpg"\\
\}
}
\end{tcolorbox}

\begin{tcolorbox}[
colback=lightProxPink!10,
colframe=lightProxPink,
left=2mm, right=2mm, top=1mm, bottom=1mm,
before skip=0pt, after skip=1pt,
title=\textcolor{black}{Tool Response}]
{\small
\includegraphics[width=0.5\linewidth, keepaspectratio]{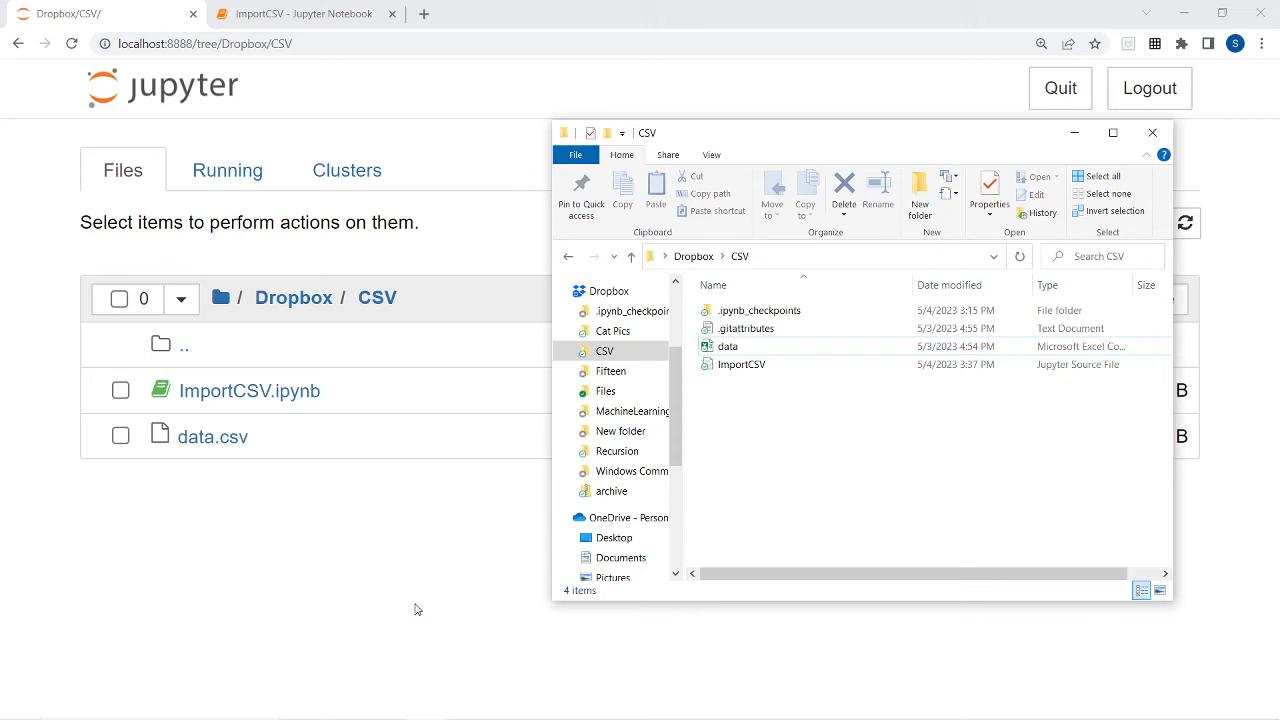}
\vfill 
}
\end{tcolorbox}

\begin{tcolorbox}[
colback=lightProxYellow!10,
colframe=lightProxYellow,
left=2mm, right=2mm, top=1mm, bottom=1mm,
before skip=0pt, after skip=1pt,
title=\textcolor{black}{read\_media\_file}]
{\small
\{\\
  "path": "frames/sec\_0010.jpg"\\
\}
}
\end{tcolorbox}

\begin{tcolorbox}[
colback=lightProxPink!10,
colframe=lightProxPink,
left=2mm, right=2mm, top=1mm, bottom=1mm,
before skip=0pt, after skip=1pt,
title=\textcolor{black}{Tool Response}]
{\small
\includegraphics[width=0.5\linewidth, keepaspectratio]{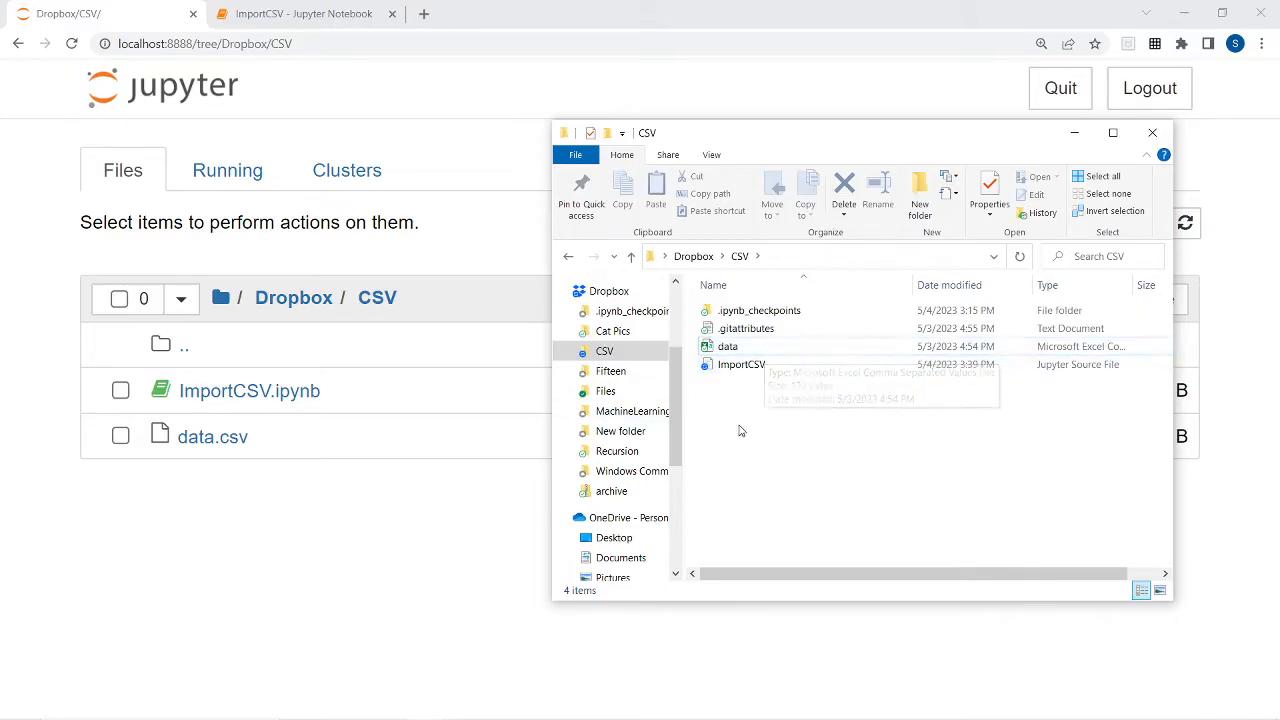}
\vfill 
}
\end{tcolorbox}

\begin{tcolorbox}[
colback=lightProxGreen!10,
colframe=lightProxGreen,
left=2mm, right=2mm, top=1mm, bottom=1mm,
before skip=0pt, after skip=1pt,
title=\textcolor{black}{Thinking}]
{\small
I can see the video content now. Let me analyze the frames:\\
\\
1. First image (sec\_0000.jpg): Shows the file explorer with "data.csv" and "Untitled.ipynb" files\\
2. Second image (sec\_0010.jpg): Shows the Jupyter notebook with:\\
   - "import pandas" \\
   - "pandas.read\_csv('data.csv')"\\
   - "df = pandas.read\_csv('data.csv')"\\
   - "df" \\
   - The output shows a DataFrame with "kind" column containing "cat" and "dog" values\\
\\
Let me continue viewing more frames to find the key frames for Data Preparation and Code Example sections.
}
\end{tcolorbox}

\begin{tcolorbox}[
colback=lightProxYellow!10,
colframe=lightProxYellow,
left=2mm, right=2mm, top=1mm, bottom=1mm,
before skip=0pt, after skip=1pt,
title=\textcolor{black}{read\_media\_file}]
{\small
\{\\
  "path": "frames/sec\_0020.jpg"\\
\}
}
\end{tcolorbox}

\begin{tcolorbox}[
colback=lightProxPink!10,
colframe=lightProxPink,
left=2mm, right=2mm, top=1mm, bottom=1mm,
before skip=0pt, after skip=1pt,
title=\textcolor{black}{Tool Response}]
{\small
\includegraphics[width=0.5\linewidth, keepaspectratio]{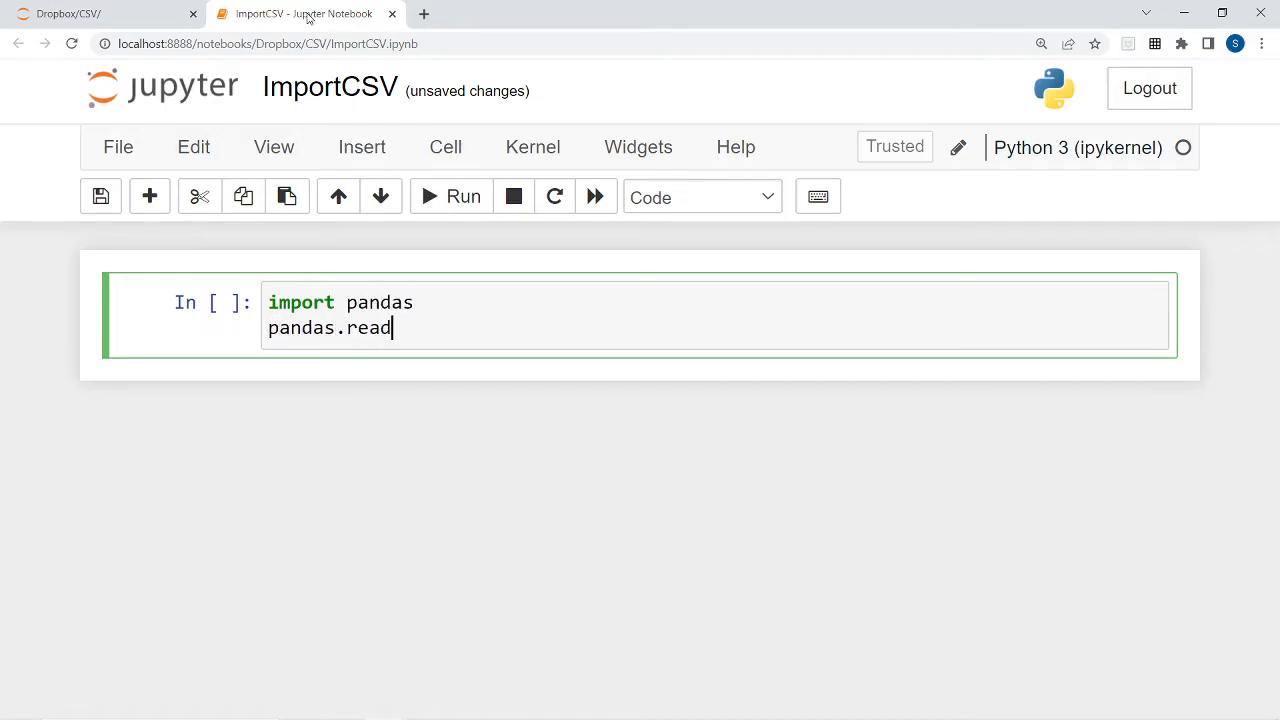}
\vfill 
}
\end{tcolorbox}

\begin{tcolorbox}[
colback=lightProxYellow!10,
colframe=lightProxYellow,
left=2mm, right=2mm, top=1mm, bottom=1mm,
before skip=0pt, after skip=1pt,
title=\textcolor{black}{read\_media\_file}]
{\small
\{\\
  "path": "frames/sec\_0030.jpg"\\
\}
}
\end{tcolorbox}

\begin{tcolorbox}[
colback=lightProxPink!10,
colframe=lightProxPink,
left=2mm, right=2mm, top=1mm, bottom=1mm,
before skip=0pt, after skip=1pt,
title=\textcolor{black}{Tool Response}]
{\small
\includegraphics[width=0.5\linewidth, keepaspectratio]{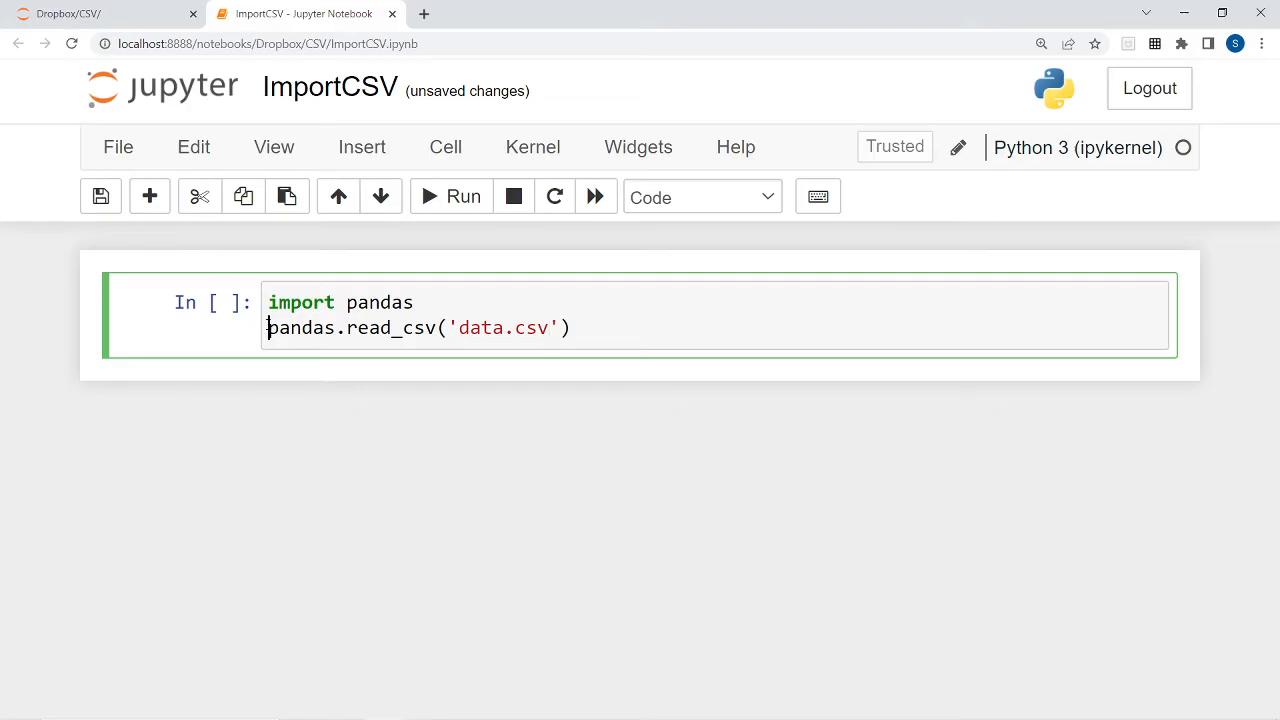}
\vfill 
}
\end{tcolorbox}

\begin{tcolorbox}[
colback=lightProxGreen!10,
colframe=lightProxGreen,
left=2mm, right=2mm, top=1mm, bottom=1mm,
before skip=0pt, after skip=1pt,
title=\textcolor{black}{Thinking}]
{\small
Based on the video frames I've seen, I can now understand the content:\\
\\
1. **Data Preparation (sec\_0000 and sec\_0010 frames)**:\\
   - Shows a folder with `animals.csv` and `Jupyter Notebook.ipynb`\\
   - The instruction is to put the CSV file in the same folder as the Jupyter notebook\\
\\
2. **Code Example (sec\_0020 and sec\_0030 frames)**:\\
   - Shows the Jupyter notebook with the code:\\
     - `import pandas`\\
     - `df = pandas.read\_csv("animals.csv")`\\
     - `df`\\
   - Shows the result with cat and dog data (the audio mentions "cats and dogs")\\
\\
Now I need to:\\
1. Create the `assets` folder\\
2. Capture the key frames for each part:\\
   - `data\_preparation.png` - should show the folder with CSV file and notebook\\
   - `code\_example.png` - should show the Jupyter notebook code and result\\
3. Create the `tutorial.md` file\\
\\
Let me identify the best frames:\\
- For Data Preparation: sec\_0000 or sec\_0010 - they show the folder with CSV file\\
- For Code Example: sec\_0020 or sec\_0030 - they show the code and result\\
\\
Let me capture these specific frames and save them to the assets folder.
}
\end{tcolorbox}

\begin{tcolorbox}[
colback=lightProxYellow!10,
colframe=lightProxYellow,
left=2mm, right=2mm, top=1mm, bottom=1mm,
before skip=0pt, after skip=1pt,
title=\textcolor{black}{create\_directory}]
{\small
\{\\
  "path": "assets"\\
\}
}
\end{tcolorbox}

\begin{tcolorbox}[
colback=lightProxPink!10,
colframe=lightProxPink,
left=2mm, right=2mm, top=1mm, bottom=1mm,
before skip=0pt, after skip=1pt,
title=\textcolor{black}{Tool Response}]
{\small
Successfully created directory assets
}
\end{tcolorbox}

\begin{tcolorbox}[
colback=lightProxGreen!10,
colframe=lightProxGreen,
left=2mm, right=2mm, top=1mm, bottom=1mm,
before skip=0pt, after skip=1pt,
title=\textcolor{black}{Thinking}]
{\small
Now let me capture the key frames for each part. I'll use the frames I already exported. Let me save them with the appropriate names.\\
\\
For Data Preparation, I'll use sec\_0000.jpg (shows the folder with animals.csv)\\
For Code Example, I'll use sec\_0020.jpg or sec\_0030.jpg (shows the code and result)\\
\\
Actually, I should export specific frames. Let me check the frame at around 15 seconds for a clearer code example view, and save the appropriate frames.
}
\end{tcolorbox}

\begin{tcolorbox}[
colback=lightProxYellow!10,
colframe=lightProxYellow,
left=2mm, right=2mm, top=1mm, bottom=1mm,
before skip=0pt, after skip=1pt,
title=\textcolor{black}{export\_frame\_at\_second}]
{\small
\{\\
  "output\_path": "assets/data\_preparation.png",\\
  "second": 2,\\
  "video\_path": "Notebook.mp4"\\
\}
}
\end{tcolorbox}

\begin{tcolorbox}[
colback=lightProxPink!10,
colframe=lightProxPink,
left=2mm, right=2mm, top=1mm, bottom=1mm,
before skip=0pt, after skip=1pt,
title=\textcolor{black}{Tool Response}]
{\small
assets/data\_preparation.png
}
\end{tcolorbox}

\begin{tcolorbox}[
colback=lightProxYellow!10,
colframe=lightProxYellow,
left=2mm, right=2mm, top=1mm, bottom=1mm,
before skip=0pt, after skip=1pt,
title=\textcolor{black}{export\_frame\_at\_second}]
{\small
\{\\
  "output\_path": "assets/code\_example.png",\\
  "second": 22,\\
  "video\_path": "Notebook.mp4"\\
\}
}
\end{tcolorbox}

\begin{tcolorbox}[
colback=lightProxPink!10,
colframe=lightProxPink,
left=2mm, right=2mm, top=1mm, bottom=1mm,
before skip=0pt, after skip=1pt,
title=\textcolor{black}{Tool Response}]
{\small
assets/code\_example.png
}
\end{tcolorbox}

\begin{tcolorbox}[
colback=lightProxGreen!10,
colframe=lightProxGreen,
left=2mm, right=2mm, top=1mm, bottom=1mm,
before skip=0pt, after skip=1pt,
title=\textcolor{black}{Thinking}]
{\small
Now let me verify the captured images are correct by reading them.
}
\end{tcolorbox}

\begin{tcolorbox}[
colback=lightProxYellow!10,
colframe=lightProxYellow,
left=2mm, right=2mm, top=1mm, bottom=1mm,
before skip=0pt, after skip=1pt,
title=\textcolor{black}{read\_media\_file}]
{\small
\{\\
  "path": "assets/data\_preparation.png"\\
\}
}
\end{tcolorbox}

\begin{tcolorbox}[
colback=lightProxPink!10,
colframe=lightProxPink,
left=2mm, right=2mm, top=1mm, bottom=1mm,
before skip=0pt, after skip=1pt,
title=\textcolor{black}{Tool Response}]
{\small
\includegraphics[width=0.5\linewidth, keepaspectratio]{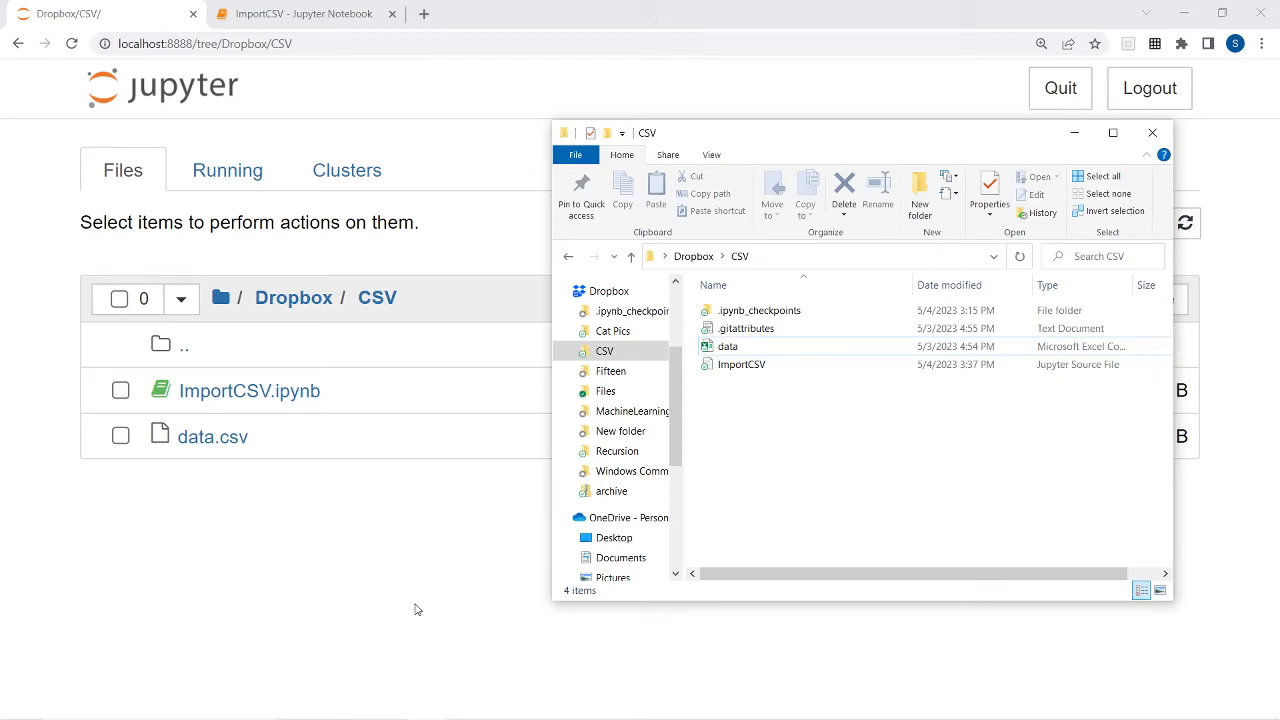}
\vfill 
}
\end{tcolorbox}

\begin{tcolorbox}[
colback=lightProxYellow!10,
colframe=lightProxYellow,
left=2mm, right=2mm, top=1mm, bottom=1mm,
before skip=0pt, after skip=1pt,
title=\textcolor{black}{read\_media\_file}]
{\small
\{\\
  "path": "assets/code\_example.png"\\
\}
}
\end{tcolorbox}

\begin{tcolorbox}[
colback=lightProxPink!10,
colframe=lightProxPink,
left=2mm, right=2mm, top=1mm, bottom=1mm,
before skip=0pt, after skip=1pt,
title=\textcolor{black}{Tool Response}]
{\small
\includegraphics[width=0.5\linewidth, keepaspectratio]{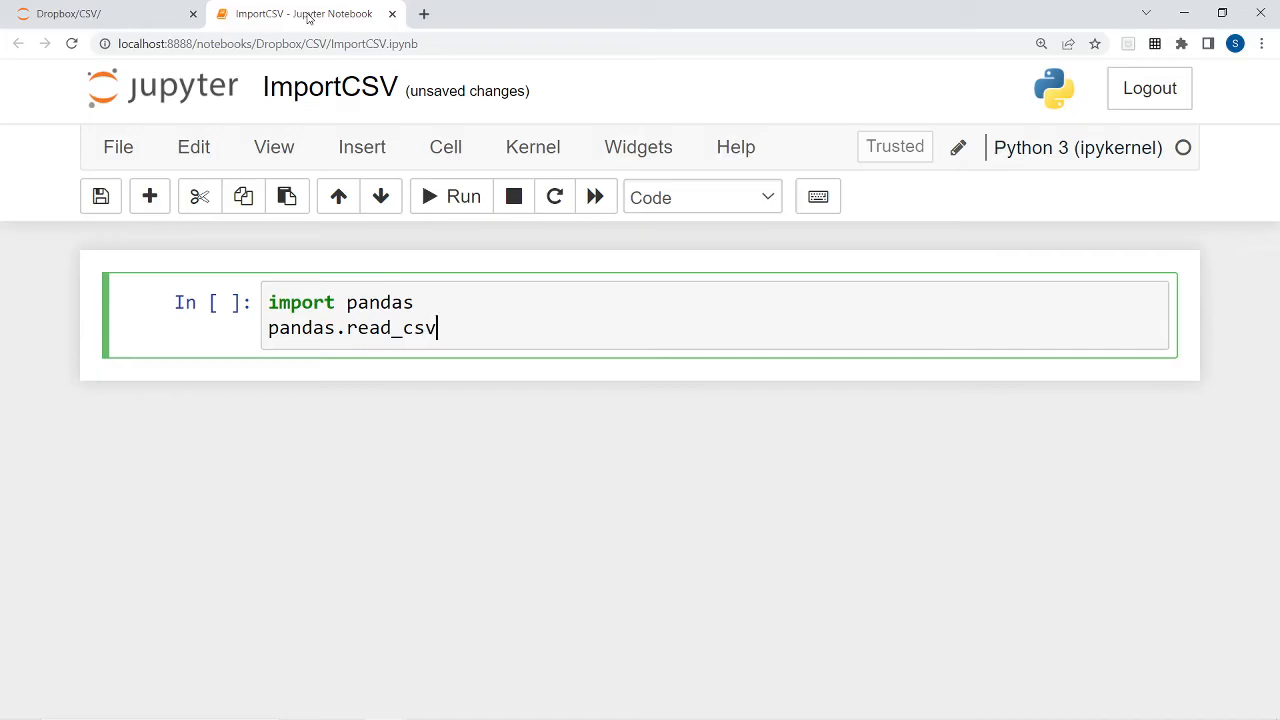}
\vfill 
}
\end{tcolorbox}

\begin{tcolorbox}[
colback=lightProxGreen!10,
colframe=lightProxGreen,
left=2mm, right=2mm, top=1mm, bottom=1mm,
before skip=0pt, after skip=1pt,
title=\textcolor{black}{Thinking}]
{\small
The images look good:\\
\\
1. **data\_preparation.png** - Shows the folder with `animals.csv` file and `Jupyter Notebook.ipynb` - perfect for the Data Preparation section\\
\\
2. **code\_example.png** - Shows the Jupyter notebook with:\\
   - Code: `import pandas` and `pandas.read\_csv('animals.csv')` and `dataframe = pandas.read\_csv('animals.csv')` and `dataframe`\\
   - Output: A table showing data about cats and dogs with columns like `id`, `name`, `type`, `age`, `weight`\\
\\
Now I can create the tutorial.md file with:\\
1. A title for the tutorial\\
2. \#\# Data Preparation section with description and image\\
3. \#\# Code Example section with description and image\\
\\
Let me write the markdown file now.
}
\end{tcolorbox}

\begin{tcblisting}{
colback=lightProxYellow!10,
colframe=lightProxYellow,
left=2mm, right=2mm, top=1mm, bottom=1mm,
before skip=0pt, after skip=1pt,
title={\textcolor{black}{write\_file}},
listing only,
breakable,
listing options={
  basicstyle=\small\ttfamily,
  breaklines=true,
  columns=fullflexible,
  keepspaces=true,
  showstringspaces=false
}
}
{
  "content": "# Pandas Tutorial: Reading CSV Data in Jupyter Notebook\n\nThis tutorial demonstrates how to use the pandas package to read a CSV file into a Jupyter notebook.\n\n## Data Preparation\n\nBefore writing code, you need to prepare your data files properly:\n\n1. **Place your CSV file in the same folder as your Jupyter notebook**: Ensure that your data file (e.g., `animals.csv`) is located in the same directory as your `.ipynb` notebook file. This makes it easy to reference the file without specifying a full path.\n\n2. **Verify the file location**: Check that both your data file and notebook are visible in the same folder directory.\n\n![Data Preparation](./assets/data_preparation.png)\n\n## Code Example\n\nFollow these steps to import your CSV data into pandas:\n\n1. **Import pandas**: Start by importing the pandas library using the import statement:\n   ```python\n   import pandas\n   ```\n\n2. **Read the CSV file**: Use `pandas.read_csv()` function with the filename in quotes:\n   ```python\n   pandas.read_csv('animals.csv')\n   ```\n\n3. **Assign to a variable**: Store the data in a DataFrame variable for easy access:\n   ```python\n   dataframe = pandas.read_csv('animals.csv')\n   ```\n\n4. **Display the data**: Simply type the variable name and run to view the contents:\n   ```python\n   dataframe\n   ```\n\n![Code Example](./assets/code_example.png)\n\nThe output will display your CSV data in a formatted table with rows and columns, making it easy to analyze and work with.\n",
  "path": "tutorial.md"
}
\end{tcblisting}

\begin{tcolorbox}[
colback=lightProxPink!10,
colframe=lightProxPink,
left=2mm, right=2mm, top=1mm, bottom=1mm,
before skip=0pt, after skip=1pt,
title=\textcolor{black}{Tool Response}]
{\small
Successfully wrote to tutorial.md
}
\end{tcolorbox}

\begin{tcolorbox}[
colback=lightProxGreen!10,
colframe=lightProxGreen,
left=2mm, right=2mm, top=1mm, bottom=1mm,
before skip=0pt, after skip=1pt,
title=\textcolor{black}{Thinking}]
{\small
The tutorial.md file has been created. Let me verify the content to make sure everything is correct.
}
\end{tcolorbox}

\begin{tcolorbox}[
colback=lightProxYellow!10,
colframe=lightProxYellow,
left=2mm, right=2mm, top=1mm, bottom=1mm,
before skip=0pt, after skip=1pt,
title=\textcolor{black}{read\_file}]
{\small
\{\\
  "path": "tutorial.md"\\
\}
}
\end{tcolorbox}

\begin{tcolorbox}[
colback=lightProxPink!10,
colframe=lightProxPink,
left=2mm, right=2mm, top=1mm, bottom=1mm,
before skip=0pt, after skip=1pt,
title=\textcolor{black}{Tool Response}]
{\small
\# Pandas Tutorial: Reading CSV Data in Jupyter Notebook\\
\\
This tutorial demonstrates how to use the pandas package to read a CSV file into a Jupyter notebook.\\
\\
\#\# Data Preparation\\
\\
Before writing code, you need to prepare your data files properly:\\
\\
1. **Place your CSV file in the same folder as your Jupyter notebook**: Ensure that your data file (e.g., `animals.csv`) is located in the same directory as your `.ipynb` notebook file. This makes it easy to reference the file without specifying a full... (truncated)
}
\end{tcolorbox}

\begin{tcolorbox}[
colback=lightProxGreen!10,
colframe=lightProxGreen,
left=2mm, right=2mm, top=1mm, bottom=1mm,
before skip=0pt, after skip=1pt,
title=\textcolor{black}{Thinking}]
{\small
The task is complete. Let me verify:\\
\\
1. I've listened to the video and understood the content\\
2. Created two main parts in the tutorial.md:\\
   - \#\# Data Preparation - describes how to prepare the data files\\
   - \#\# Code Example - describes the code steps to read CSV with pandas\\
3. Captured the key frames:\\
   - `data\_preparation.png` for Data Preparation (shows the folder with animals.csv and Jupyter Notebook)\\
   - `code\_example.png` for Code Example (shows the Jupyter notebook with code and output)\\
4. Saved images in the `assets` folder\\
5. Inserted images into the markdown document using Markdown syntax\\
\\
The tutorial.md file is complete and meets all requirements.
}
\end{tcolorbox}

\begin{tcolorbox}[
colback=lightProxYellow!10,
colframe=lightProxYellow,
left=2mm, right=2mm, top=1mm, bottom=1mm,
before skip=0pt, after skip=1pt,
title=\textcolor{black}{local-claim\_done}]
{\small
}
\end{tcolorbox}

\begin{tcolorbox}[
colback=lightProxPink!10,
colframe=lightProxPink,
left=2mm, right=2mm, top=1mm, bottom=1mm,
before skip=0pt, after skip=1pt,
title=\textcolor{black}{Tool Response}]
{\small
you have claimed the task is done!
}
\end{tcolorbox}
\end{document}